\RequirePackage{fix-cm}
\documentclass{article}
\newif\ifpreflutarxiv
\preflutarxivtrue
\usepackage{iclr2027_conference,times}

\usepackage{amsmath,amsfonts,bm}

\def\eqref#1{equation~\ref{#1}}

\def\1{\bm{1}}

\DeclareMathAlphabet{\mathsfit}{\encodingdefault}{\sfdefault}{m}{sl}
\SetMathAlphabet{\mathsfit}{bold}{\encodingdefault}{\sfdefault}{bx}{n}

\usepackage{amsmath,amssymb}
\usepackage{booktabs,longtable}
\usepackage[table]{xcolor}
\definecolor{SectionTint}{HTML}{EAF0F5}
\definecolor{WarmSection}{HTML}{F3ECE3}
\definecolor{PrefTint}{HTML}{F3F6F9}
\definecolor{PassTint}{HTML}{F6EEE5}
\definecolor{ResultInk}{HTML}{5C3834}
\definecolor{PassInk}{HTML}{705B47}
\definecolor{RefineAA}{HTML}{FBF8EE}
\definecolor{RefineAB}{HTML}{FBF8EE}
\definecolor{RefineAC}{HTML}{FBF8EE}
\definecolor{RefineAD}{HTML}{FBF8EE}
\definecolor{RefineBA}{HTML}{F6EBDC}
\definecolor{RefineBB}{HTML}{F6EBDC}
\definecolor{RefineBC}{HTML}{F7EDDF}
\definecolor{RefineBD}{HTML}{F6ECDD}
\definecolor{RefineCA}{HTML}{F2E2DA}
\definecolor{RefineCB}{HTML}{F2E1D9}
\definecolor{RefineCC}{HTML}{F3E4DA}
\definecolor{RefineCD}{HTML}{F1E0D9}
\definecolor{RefineDA}{HTML}{F1DFD9}
\definecolor{RefineDB}{HTML}{F0DED9}
\definecolor{RefineDC}{HTML}{F1E0D9}
\definecolor{RefineDD}{HTML}{F0DDD9}
\definecolor{RefineEA}{HTML}{EFDBD8}
\definecolor{RefineEB}{HTML}{EFDBD8}
\definecolor{RefineEC}{HTML}{EFDBD8}
\definecolor{RefineED}{HTML}{EFDBD8}

\newcolumntype{R}[1]{>{\raggedleft\arraybackslash}p{#1}}
\newcolumntype{L}[1]{>{\raggedright\arraybackslash}p{#1}}

\usepackage{graphicx}
\usepackage{multirow}
\usepackage{microtype}
\usepackage{hyperref}
\hypersetup{colorlinks=true,allcolors=black}
\usepackage{url}

\title{PrefLUT: Reusable and Refinable Personalized Color Editing from Pairwise Preferences}

\ifpreflutarxiv
  \iclrfinalcopy
  \author{%
    \hspace*{-\tabcolsep}\begin{minipage}[t]{\textwidth}\centering\bfseries
    Chuanzhi Xu\textsuperscript{1,*}\quad
    Langyi Chen\textsuperscript{1}\quad
    Chengkun Yue\textsuperscript{2}\quad
    Xuanhua Yin\textsuperscript{1}\\
    \bfseries Boyu Wei\textsuperscript{1}\quad
    Qingwen Zeng\textsuperscript{1}\quad
    Zihan Deng\textsuperscript{3}\quad
    Weidong Cai\textsuperscript{1}\\[4pt]
    {\normalfont\small\textsuperscript{1}The University of Sydney\quad
    \textsuperscript{2}Indiana University\quad
    \textsuperscript{3}The University of Hong Kong}\\[2pt]
    {\normalfont\small\textsuperscript{*}Corresponding author:
    \href{mailto:chuanzhi.xu@sydney.edu.au}{\texttt{chuanzhi.xu@sydney.edu.au}}\qquad
    \href{https://github.com/LouckXu/PrefLUT}{\raisebox{-1pt}{\includegraphics[height=9pt]{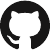}}\enspace\textcolor{blue!55!black}{\textbf{Project Page}}}}\end{minipage}\hspace*{-\tabcolsep}}
  \hypersetup{pdfauthor={Chuanzhi Xu, Langyi Chen, Chengkun Yue, Xuanhua Yin, Boyu Wei, Qingwen Zeng, Zihan Deng, Weidong Cai}}
\else
  \author{Anonymous Authors}
  \hypersetup{pdfauthor={}}
\fi
\hypersetup{pdftitle={PrefLUT: Reusable and Refinable Personalized Color Editing from Pairwise Preferences}}

\begin{document}

\maketitle
\ifpreflutarxiv
  \lhead{Preprint}
\fi
\suppressfloats[t] 

\begin{abstract}
Photographic color editing is inherently personal: the same image can appear too warm,
too muted, or already satisfactory to different users. Most lookup table (LUT) and reference-guided
methods target a specified appearance rather than model persistent preferences from
repeated user choices. To address this gap, we introduce
PrefLUT, a reusable and refinable user-preference modeling framework
for deployable 3D LUTs, encoding ordered preferred/non-preferred
image pairs into a lightweight Reusable User Profile that is reused across queries
and refined using additional user preference pairs, without per-user optimization. A
Query-Conditioned LUT Predictor combines this
profile with each image to predict a LUT latent vector and edit strength. An
Identity-Residual LUT Decoder and Edit-Strength Controller then produce an
exportable 3D LUT. Experiments on three datasets demonstrate effective personalized
editing and general-purpose enhancement. Each quantized profile requires only 260 bytes,
and editing takes 1.365\,ms/image on an RTX~5090 GPU.
We also introduce the Preference-Conditioning Verification Protocol (PCVP), an
evaluation protocol to verify whether personalized image edits depend on user preferences
and the query image through controlled changes to user profiles, preference orders,
pair correspondences, and query images.
\end{abstract}

\section{Introduction}
\label{sec:introduction}

Color and tone shape a photograph's perceived quality, mood, and visual
intent. Modern retouching methods learn fast global or spatially aware color
transforms from expert-edited images
\citep{bychkovsky2011fivek,zeng2022lut,yang2022adaint,yang2022seplut,kim2024bgrid}.
Lookup table (LUT) models predict lightweight color mappings that can be applied to
full-resolution images, exported, inspected, and reused, making LUTs attractive for
efficient color editing.
\begin{figure*}[t]
  \centering
  \includegraphics[width=\textwidth]{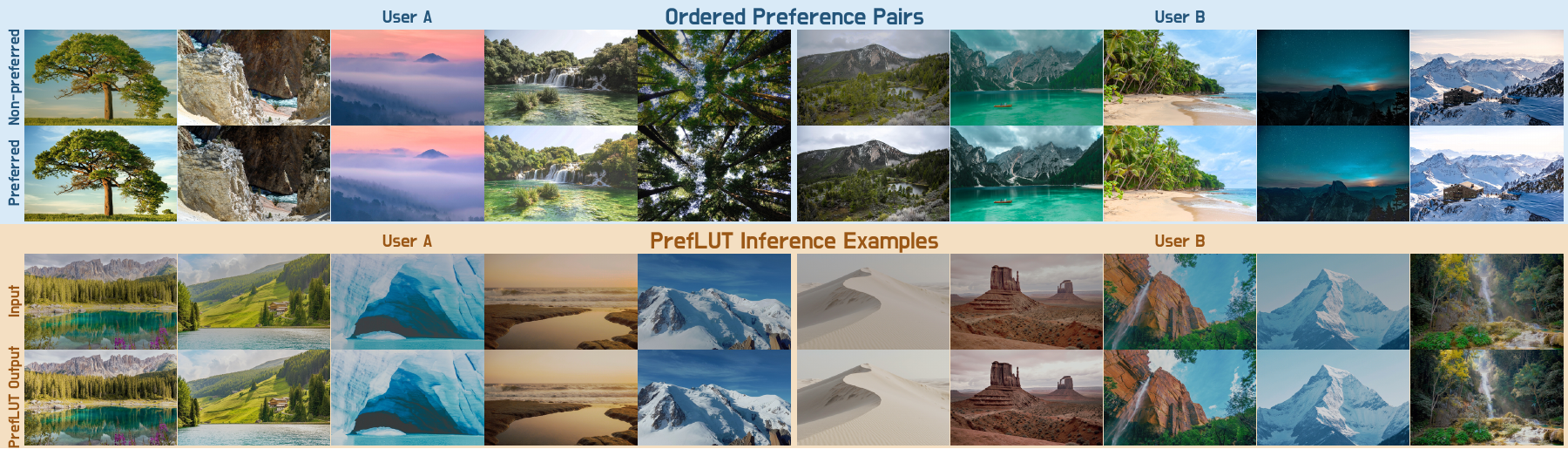}
  \caption{Preference-conditioned editing across users. Examples from two
  PPSD users show ordered preferred/non-preferred image pairs (top) and unseen inputs
  with PrefLUT outputs (bottom). PrefLUT outputs reflect each user's color
  preferences across different scenes.}
  \label{fig:teaser}
\end{figure*}
Nevertheless, most learned-LUT methods optimize a shared target, such as one retoucher in
MIT-Adobe FiveK or one expert in PPR10K
\citep{bychkovsky2011fivek,liang2021ppr10k}, whereas reference-guided methods use a
single style image to define the desired appearance and predict a preset, implicit color map,
3D LUT, or spatial 4D LUT
\citep{ho2021deeppreset,lin2023adacm,ke2023neuralpreset,li2025dlut,gong2025salut,
ma2026acetone}.

However, a personal preference is different: it is expressed through repeated
choices, may be stable only along selected color dimensions, must be separated
from image content unrelated to preferences, and refined as feedback accumulates.
The Personalized Photographic Style Dataset (PPSD) establishes this setting through pairwise user choices and evaluates whether an
output remains faithful while moving toward a user's preferred appearance
\citep{kim2026ppsd}. However, its proposed baselines do not explicitly output LUTs.
Moreover, without comparing edits of the same query conditioned on the intended user,
another user, or a population-level preference, its preference-alignment scores alone
cannot distinguish the use of user-specific preferences from editing tendencies shared across users.

This gap raises a central question: \emph{Can preferred/non-preferred image pairs form a reusable
user profile that can be refined by aggregating new feedback again to predict
explicit LUTs for new images without per-user optimization, and can we verify that these
edits reflect the user's preferences and each query image's content rather than editing tendencies shared across users?}

We answer this question with PrefLUT, which is a reusable and refinable
user-preference modeling framework designed for deployable 3D LUTs. The framework
divides personalized color editing into profile construction and query-time editing.
During profile construction, an Ordered Preference Encoder maps each ordered
preference pair to an ordered preference token, and the Preference Set
Aggregator summarizes the tokens before a Reusable User Profile is stored.
During query-time editing, a Query Image Encoder extracts a query feature from
the new image. The Query-Conditioned LUT Predictor combines this feature with the
Reusable User Profile to predict a LUT latent vector and edit strength. An Identity-Residual LUT Decoder converts
the latent vector into a LUT, and the Edit-Strength Controller scales its residual before
full-resolution application. New preference selections or user-confirmed edits provide
additional preferred/non-preferred image pairs, which are aggregated with retained pairs
to refine the profile while keeping network weights frozen.
Figure~\ref{fig:teaser} shows examples of personalized color editing \mbox{for different users.}

We evaluate PrefLUT on three datasets and introduce the Preference-Conditioning
Verification Protocol (PCVP) to test whether edits respond to user preferences and
the query image. Results demonstrate strong personalized editing
and general-purpose enhancement. Experiments that accumulate retouching preference feedback to refine
profiles show improved editing on unseen queries without updating network
weights. Our contributions can be summarized as follows:
\begin{itemize}
  \item We propose PrefLUT, a reusable and refinable user-preference modeling framework
  for deployable 3D LUTs. Its profile can be rebuilt from updated feedback and supports explicit,
  query-conditioned LUT prediction for new images without per-user optimization.
  \item We introduce PCVP as a systematic verification protocol for personalized image
  enhancement and color grading. Its five controlled tests assess quality gains from the
  intended user's preferences and current query image, providing the field with
  a rigorous personalization test.
  \item Experiments on three datasets demonstrate effective personalized editing and
  general-purpose enhancement. PrefLUT leads all four direct fidelity metrics and
  achieves the highest Comparative Quality Score (CQS)~\citep{kim2026ppsd} under $\Delta E_{00}$, LPIPS, and SSIM among
  the compared PPSD methods and passes all five
  PCVP tests. Additional preference
  feedback improves editing on unseen queries through profile refinement with frozen
  network weights.
\end{itemize}

\section{Related Work}

\noindent\textbf{Learned LUTs and Reusable Color Transforms.}\quad
Learned LUT methods express global or spatial color edits as lightweight mappings without
requiring full-resolution neural decoding. LUT-based representatives include Image-Adaptive 3D-LUT,
Spatial-Aware 3D LUT, 4D LUT, BGrid, AdaInt, SepLUT, CLUT-Net, NILUT, and neural LUT encoding
\citep{zeng2022lut,wang2021s3dlut,liu2023fourdlut,kim2024bgrid,yang2022adaint,
yang2022seplut,zhang2022clut,conde2024nilut,zehtab2025encoding}. Meanwhile,
preset-based methods such as Deep Preset, AdaCM, and Neural Preset encode reusable
appearance through retouching parameters, implicit mappings, or style representations
\citep{ho2021deeppreset,lin2023adacm,ke2023neuralpreset}. However, these methods
are typically trained to reproduce a shared target, select from a fixed set of styles,
or transfer the appearance of a single reference image. In contrast, our PrefLUT infers a persistent preference from multiple
ordered choices and predicts an explicit query-conditioned LUT, enabling reusable and refinable
personalization, efficient full-resolution application, and direct export \mbox{to color-grading workflows.}

\noindent\textbf{Reference-Guided and Generative LUTs.}\quad
Reference-image-guided and generative LUT methods use a current image, reference, text,
or generative prior to produce flexible color transforms. Representative methods include
D-LUT, SA-LUT, AceTone, StatLUT, and FlowLUT
\citep{li2025dlut,gong2025salut,ma2026acetone,wang2026statlut,hu2025flowlut}.
However, they condition each edit independently and do not aggregate repeated ordered
choices into a persistent user preference. In contrast, our PrefLUT updates the profile
when feedback arrives and reuses it to predict explicit query-conditioned LUTs,
without reprocessing preference references at query time.

\noindent\textbf{Personalized Aesthetic Enhancement.}\quad
Personalized enhancement follows individual preferences, whereas
general-purpose enhancement optimizes a shared aesthetic target.
PieNet encodes preferred images into user preference vectors
\citep{kim2020pienet}. StarEnhancer represents multiple tonal styles within one model
\citep{song2021starenhancer}, while PIE-MSM uses masked style modeling to predict
content-aware edits from preferred images \citep{kosugi2023masked}.
PerTouch combines diffusion-based retouching with user
feedback and scene-aware memory \citep{chang2026pertouch}.
PPSD evaluates User-specific Decoder, User Preference Embedding (UPE), and
Exemplar-based Inference for learning pairwise preferences
\citep{kim2026ppsd}. However, these methods do not jointly support profile refinement
from pairwise feedback and exportable LUT-based editing. PrefLUT uses
a compact, reusable profile refined from preference pairs to predict
exportable, query-conditioned 3D LUTs.

\section{PrefLUT}

\begin{figure*}[t]
  \centering
  \includegraphics[width=\textwidth]{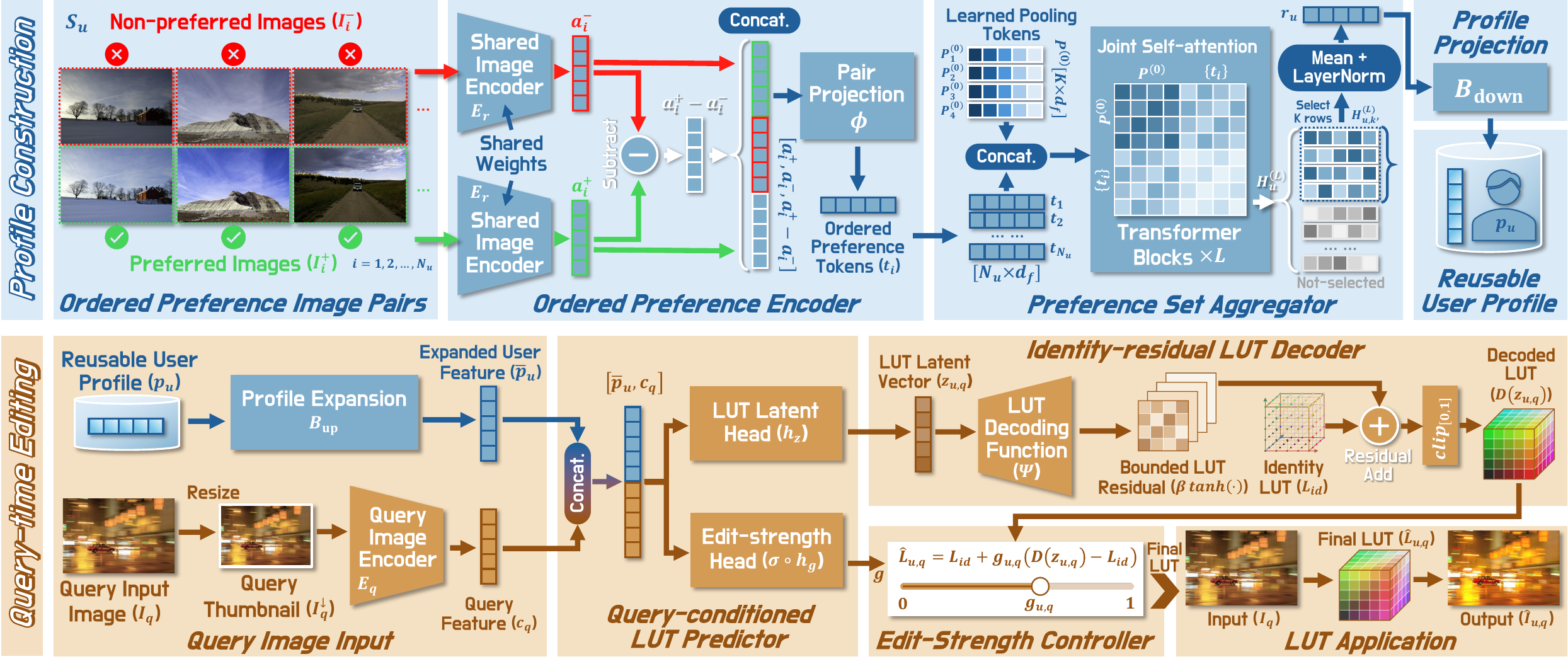}
  \caption{PrefLUT pipeline. Profile Construction (top) encodes ordered preference
  image pairs into a Reusable User Profile that can be refined with new feedback.
  Query-Time Editing (bottom) combines the stored profile with each query image to predict
  a 3D LUT for full-resolution editing.}
  \label{fig:overview}
\end{figure*}

\noindent\textbf{Overview and Problem Formulation.}\quad
Figure~\ref{fig:overview} illustrates the two stages of PrefLUT: Profile Construction
and Query-Time Editing. For user $u$, the current reference set
$S_u=\{(I_i^{+},I_i^{-})\}_{i=1}^{N_u}$ contains $N_u$ ordered preference image pairs used
to construct the profile, where $I_i^{+}$ and $I_i^{-}$ are the preferred and
non-preferred images for user $u$. Given an unseen
query image $I_q$ and its resized
thumbnail $I_q^{\downarrow}$, PrefLUT predicts a global 3D LUT $\widehat L_{u,q}$. Its
trilinear color mapping $\mathcal T(\widehat L_{u,q},I_q)$ should move the query toward
the user's repeated color choices in $S_u$ while preserving its content.

The pipeline outlined in Section~\ref{sec:introduction} and Appendix~\ref{app:method-data-flow}
is expressed mathematically as:
\begin{equation}
  \begin{alignedat}{2}
  p_u &= f_{\mathrm{profile}}(S_u),\qquad &
  (z_{u,q},g_{u,q}) &= f_{\mathrm{query}}(p_u,I_q^{\downarrow}),\\
  \widehat L_{u,q} &= f_{\mathrm{LUT}}(z_{u,q},g_{u,q}),\qquad &
  \widehat I_{u,q} &= \mathcal T(\widehat L_{u,q}, I_q).
  \end{alignedat}
  \label{eq:factorization}
\end{equation}
Here $f_{\mathrm{profile}}$ constructs or refines the profile from the current $S_u$, $f_{\mathrm{query}}$ predicts the
LUT latent vector and edit strength, and $f_{\mathrm{LUT}}$ produces the final LUT.
$\widehat I_{u,q}$ is the edited image. $p_u\in\mathbb{R}^{d_p}$ is the
Reusable User Profile, where $d_p$ is the profile width.
$z_{u,q}\in\mathbb{R}^{d_z}$ is the lightweight LUT latent vector predicted for
the current query, where $d_z$ is the LUT latent vector width. Finally,
$g_{u,q}\in[0,1]$ is its edit strength.

\subsection{Profile Construction}

\noindent\textbf{Ordered Preference Encoder.}\quad
Each image in an ordered preference image pair passes through the same reference image
encoder $E_r$. For pair $i$, the preferred and non-preferred features are
$a_i^+=E_r(I_i^+)$ and $a_i^-=E_r(I_i^-)$. The Ordered Preference Encoder
keeps the preference direction in each token:
\begin{equation}
  t_i=\phi\!\left([a_i^+,a_i^-,a_i^+-a_i^-]\right)\in\mathbb R^{d_f},
  \label{eq:pairtoken}
\end{equation}
where $d_f$ is the feature width, $[\cdot]$ denotes concatenation, and $\phi$ is a
learned pair projection. The first
two terms retain the preferred and non-preferred image features, while
$a_i^+-a_i^-$ records the feature change from the non-preferred image to the preferred
image. Swapping the preference labels reverses both the feature
order and the sign of the difference. $t_i$ is termed an ordered preference token. It
records preference direction without requiring pixel alignment.

\noindent\textbf{Preference Set Aggregator and Reusable User Profile.}\quad
The Preference Set Aggregator combines all ordered preference tokens into
a user feature that is invariant to their list order. The aggregator prepends $K$ learned pooling tokens
$P^{(0)}\in\mathbb R^{K\times d_f}$ to the ordered preference tokens and processes
them jointly with $L$ standard Transformer blocks that do not use positional embeddings,
following the attention-based set formulation of Set Transformer
\citep{lee2019settransformer}:
\begin{equation}
  H_u^{(0)}=[P^{(0)},t_1,\ldots,t_{N_u}],\qquad
  H_u^{(\ell+1)}=\operatorname{TrBlock}_{\ell}(H_u^{(\ell)}).
  \label{eq:setaggregation}
\end{equation}
Here $H_u^{(\ell)}$ is the token sequence after $\ell$ blocks, and
$\operatorname{TrBlock}_{\ell}$ is the $\ell$-th Transformer block.
Appendix~\ref{app:batching} details batching and valid-pair masking.
After the final block, the learned pooling token outputs are averaged and layer-normalized to
obtain the aggregated user feature
$r_u=\operatorname{LN}(K^{-1}\sum_{k=1}^{K}H_{u,k}^{(L)})\in\mathbb R^{d_f}$,
where $k$ indexes the $K$ learned pooling tokens. The Profile Projection
$B_{\mathrm{down}}$ then produces the Reusable User Profile
$p_u=B_{\mathrm{down}}(r_u)\in\mathbb R^{d_p}$.
The profile is the only per-user data needed at query time. When users make new choices
or confirm manual edits, this feedback is converted into ordered preference image pairs
to refine the profile. Selected and rejected images, or edited results and preceding
candidates, serve as the preferred and non-preferred images, respectively.
These new pairs are incorporated into the reference set at update $t$. The retained
pairs are then re-encoded and aggregated to obtain
$p_u^{(t)}=f_{\mathrm{profile}}(S_u^{(t)})$, with all network weights frozen.
$S_u^{(t)}$ retains either cumulative feedback or a recent feedback window.
The updated profile is stored and reused to predict a personalized LUT for
\mbox{each subsequent query image.}

\subsection{Query-Time Editing}

\noindent\textbf{Query-Conditioned LUT Predictor.}\quad
At query time, Profile Expansion maps the profile back to the feature width,
while the separate Query Image Encoder extracts the query feature from
the thumbnail. The expanded user feature summarizes the
currently retained preferences, while the query feature describes the content and color
distribution of the current image. The Query-Conditioned LUT Predictor
concatenates these two vectors and passes them to the LUT latent vector head \mbox{and edit-strength head:}
\begin{equation}
  \begin{alignedat}{2}
  \bar p_u &= B_{\mathrm{up}}(p_u),\qquad & c_q &= E_q(I_q^{\downarrow}),\\
  z_{u,q} &= h_z([\bar p_u,c_q]),\qquad & g_{u,q} &= \sigma\!\left(h_g([\bar p_u,c_q])\right),
  \end{alignedat}
  \label{eq:query}
\end{equation}
where $h_z$ and $h_g$ are the LUT latent vector and edit-strength heads, and $\sigma$ is
the sigmoid function.
PrefLUT predicts a new LUT latent vector and edit strength for every query.

\noindent\textbf{Identity-Residual LUT Decoder.}\quad
PrefLUT uses a learned low-dimensional Identity-Residual LUT Decoder. For grid resolution $R$, the Identity-Residual LUT Decoder
predicts a residual around the identity LUT and clips the resulting LUT to
$[0,1]^{R\times R\times R\times3}$:
\begin{equation}
  D(z)=\operatorname{clip}_{[0,1]}\!\left(
  L_{\mathrm{id}}+\beta\tanh\!\big(\Psi(z)\big)\right),
  \label{eq:decoder}
\end{equation}
where $L_{\mathrm{id}}$ is the identity LUT, $\Psi$ is the learned LUT decoding function,
and $\beta$ limits the residual amplitude. $D(z)$ is termed the decoded LUT.
The decoder is pretrained on fitted target LUTs and frozen during personalized training,
when the losses act on edited images. Thus, preference training does not require a target
LUT for every pair. Appendices~\ref{app:target-lut-fitting} and~\ref{app:decoder-pretraining}
detail target fitting \mbox{and decoder pretraining.}

\noindent\textbf{Edit-Strength Controller and LUT Application.}\quad
$D(z_{u,q})-L_{\mathrm{id}}$ is the predicted LUT residual. The
Edit-Strength Controller scales this residual before it is applied to the query:
\begin{equation}
  \widehat L_{u,q}=L_{\mathrm{id}}+g_{u,q}
  \left(D(z_{u,q})-L_{\mathrm{id}}\right),
  \qquad
  \widehat I_{u,q}=\mathcal T(\widehat L_{u,q},I_q).
  \label{eq:gatedlut}
\end{equation}
Here $g_{u,q}$ depends on both the Reusable User Profile and the query image. $\widehat L_{u,q}$ is the final LUT. Trilinear interpolation is differentiable during training, and its
cost grows linearly with the number of output pixels. The final LUT has a fixed grid
size, so it can be saved and applied to \mbox{images at different resolutions.}

\subsection{Training with Separate Query Pairs}

For each user, a reference set $S_u$ and separate query pairs are sampled anew each
training epoch. The following equations describe one sampled query pair
$(Q_u^+,Q_u^-)$, where $Q_u^+$ is preferred to $Q_u^-$.
Here $\widehat I(p,Q)$ denotes PrefLUT's output for profile $p$ and query image $Q$.
Both query images use the same profile, yielding $Y_u^-=\widehat I(p_u,Q_u^-)$ and
$Y_u^+=\widehat I(p_u,Q_u^+)$.

Most ordered preference image pairs are not geometrically aligned, so PrefLUT uses a global
color-statistics descriptor $\chi$ that is less sensitive to spatial misalignment. It summarizes
the means and standard deviations of color, luminance, and saturation
(Appendix~\ref{app:color-statistics}). For images $A$ and $B$, the distance is
$d_\chi(A,B)=\operatorname{mean}(|\chi(A)-\chi(B)|)$.
Let $d^+$ be the distance from the edited non-preferred query to the preferred query,
$d^-$ the distance to the non-preferred query, and $d^0$ the distance between the query
pair:
$d^+=d_\chi(Y_u^-,Q_u^+)$,
$d^-=d_\chi(Y_u^-,Q_u^-)$, and
$d^0=d_\chi(Q_u^-,Q_u^+)$. The color-distance loss and ranking loss are:
\begin{equation}
  \mathcal L_{\mathrm{color}}=\frac{d^+}{d^0+\tau},\qquad
  \mathcal L_{\mathrm{rank}}=[m+d^+-d^-]_+,
  \label{eq:colorloss}
\end{equation}
where $[x]_+=\max(x,0)$, $\tau>0$ prevents instability when the distance between the
two query images is small, and $m>0$ is the ranking margin. The auxiliary losses enforce
reconstruction on aligned pairs ($\mathcal L_{\mathrm{aligned}}$), preservation of preferred
inputs ($\mathcal L_{\mathrm{preserve}}$), edit/preserve strength supervision
($\mathcal L_{\mathrm{strength}}$, with targets 1 for non-preferred queries and 0 for preferred queries), and limited LUT deviation from identity
($\mathcal L_{\mathrm{LUT}}$). Their definitions are given in Appendix~\ref{app:auxiliary-losses}.

Without an explicit comparison between users, the model could produce the same average
edit for every user even when $p_u$ contains user-specific information. The Wrong-User Contrast Loss
is introduced to address this. For the same query $Q_u^-$, $p_u$ is replaced with a
wrong-user profile $p_{u'}$, where $u'\neq u$
(Appendix~\ref{app:wrong-user-selection}). If
$\widetilde Y_{u\leftarrow u'}^-$ is the resulting output and
$\widetilde d^+=d_\chi(\widetilde Y_{u\leftarrow u'}^-,Q_u^+)$, then:
\begin{equation}
  \mathcal L_{\mathrm{user}}=[m+d^+-\widetilde d^+]_+.
  \label{eq:usercontrast}
\end{equation}
The loss is zero only when the Correct User Profile output is at least $m$ closer to
$Q_u^+$ than the Fixed Wrong-User Profile output for the same query. The complete objective is:
\begin{equation}
  \begin{aligned}
  \mathcal L={}&\lambda_c\mathcal L_{\mathrm{color}}
  +\lambda_r\mathcal L_{\mathrm{rank}}
  +\lambda_a\mathcal L_{\mathrm{aligned}}
  +\lambda_p\mathcal L_{\mathrm{preserve}}\\
  &+\lambda_g\mathcal L_{\mathrm{strength}}
  +\lambda_l\mathcal L_{\mathrm{LUT}}
  +\lambda_u\mathcal L_{\mathrm{user}}.
  \end{aligned}
  \label{eq:loss}
\end{equation}
\suppressfloats[t]
\begin{table*}[t]
  \centering
  \caption{Personalized editing quality and efficiency on PPSD. (A) Direct fidelity to preferred target images (Preferred) for edited non-preferred queries. (B) Personalized editing quality measured by CQS. (C) Model size and inference efficiency. $\dagger$ marks baselines adapted to PPSD.}
  \label{tab:adapted-personalized}
  \label{tab:personalized-efficiency-main}
  \small
  \definecolor{DirectColumnA}{HTML}{EDF5FB}
  \definecolor{DirectColumnB}{HTML}{DCEBF6}
  \definecolor{DirectHeader}{HTML}{C7DFEF}
  \definecolor{QualityColumnA}{HTML}{FBF6EF}
  \definecolor{QualityColumnB}{HTML}{F3E7D8}
  \definecolor{QualityHeader}{HTML}{F0DFC9}
  \colorlet{EfficiencyColumnA}{DirectColumnA}
  \colorlet{EfficiencyColumnB}{DirectColumnB}
  \colorlet{EfficiencyHeader}{DirectHeader}
  \setlength{\tabcolsep}{1.5pt}
  \newlength{\CQSColumnWidth}
  \settowidth{\CQSColumnWidth}{\textbf{(B) Personalized Editing Quality}}
  \setlength{\CQSColumnWidth}{\dimexpr\CQSColumnWidth/4-2\tabcolsep\relax}
  \resizebox{\textwidth}{!}{%
  \begin{tabular}{@{}l|*{2}{>{\columncolor{DirectColumnA}}r>{\columncolor{DirectColumnB}}r}|*{2}{>{\columncolor{QualityColumnA}}w{r}{\CQSColumnWidth}>{\columncolor{QualityColumnB}}w{r}{\CQSColumnWidth}}|*{2}{>{\columncolor{EfficiencyColumnA}}r>{\columncolor{EfficiencyColumnB}}r}>{\columncolor{EfficiencyColumnA}}r@{}}
    \toprule
   & \multicolumn{4}{c|}{\cellcolor{DirectHeader}\textbf{(A) Direct Fidelity to Preferred}}
   & \multicolumn{4}{c|}{\cellcolor{QualityHeader}\textbf{(B) Personalized Editing Quality}}
   & \multicolumn{5}{c}{\cellcolor{EfficiencyHeader}\textbf{(C) Model Size and Inference Efficiency}} \\
   & \multicolumn{4}{c|}{\cellcolor{DirectColumnA}Output vs.\ Target}
   & \multicolumn{4}{c|}{\cellcolor{QualityColumnA}CQS $\uparrow$}
   & Params & Compute & Profile & Construct. & Editing \\
    \cmidrule(lr){2-5}\cmidrule(lr){6-9}\cmidrule(lr){10-14}
  Method & $\Delta E_{00}\downarrow$ & LPIPS $\downarrow$ & PSNR $\uparrow$ & SSIM $\uparrow$
   & $\Delta E_{00}\uparrow$ & LPIPS $\uparrow$ & PSNR $\uparrow$ & SSIM $\uparrow$
   & (M) & (GFLOPs) & (B/user) & (ms/user) & (ms/image) \\
    \midrule
  PieNet$^{\dagger}$ & 8.432 & 0.118 & 21.719 & 0.818 & 0.218 & 14.978 & 36.611 & 0.901 & 26.349 & 76.094 & 2,048 & 130.626 & 4.091 \\
  StarEnhancer$^{\dagger}$ & 8.077 & 0.090 & 21.949 & 0.835 & 0.213 & 21.477 & 30.960 & 0.918 & 38.052 & 3.638 & 4,096 & 141.096 & 3.253 \\
  PIE-MSM$^{\dagger}$ & 8.179 & 0.092 & 21.576 & 0.829 & 0.173 & 17.711 & 26.827 & 0.905 & 90.756 & 55.909 & 32,768 & 119.766 & 9.401 \\
  DiffRetouch$^{\dagger}$ & 9.380 & 0.108 & 20.881 & 0.823 & 0.121 & 13.184 & 22.328 & 0.884 & 924.139 & 15,939.293 & \textbf{16} & 12.705 & 653.309 \\
  PerTouch$^{\dagger}$ & 8.615 & 0.115 & 21.184 & 0.795 & 0.147 & 12.301 & 23.487 & 0.850 & 1,654.939 & 58,386.024 & \textbf{16} & 76.520 & 1,765.465 \\
  User-specific Decoder & 9.493 & 0.105 & 20.612 & 0.830 & 0.166 & 18.086 & 29.414 & 0.912 & \textbf{2.956} & 1,551.523 & 6,940,284 & 3,518.416 & 70.406 \\
  UPE & 8.929 & 0.100 & 20.882 & 0.832 & 0.217 & 19.900 & \textbf{39.206} & 0.916 & 26.330 & 1,585.883 & 1,024 & 117.030 & 73.637 \\
  Exemplar-based Inference & 14.330 & 0.163 & 16.921 & 0.750 & 0.082 & 8.342 & 19.136 & 0.817 & 92.330 & 55.960 & 49,152 & 122.329 & 9.562 \\
    \midrule
  PrefLUT (ours) & \textbf{8.067} & \textbf{0.089} & \textbf{22.094} & \textbf{0.843} & \textbf{0.219} & \textbf{21.615} & 32.734 & \textbf{0.920} & 10.264 & \textbf{1.966} & 260 & \textbf{2.938} & \textbf{1.365} \\
    \bottomrule
  \end{tabular}}
\end{table*}

\section{Preference-Conditioning Verification Protocol}
\label{sec:pcvp}

Personalized image enhancement is commonly evaluated by comparing outputs with
retouched targets and measuring preference alignment \citep{kim2020pienet,kim2026ppsd}.
However, such scores alone cannot distinguish the use of individual preferences from editing
tendencies shared across users. Therefore, we believe the field needs a common evaluation protocol
to verify user-specific conditioning by comparing edits of the same query under
\mbox{intended-user, other-user, and population preferences.}

\noindent\textbf{Evaluation Principle.}\quad We introduce PCVP to verify whether a personalized image enhancement or color-grading method uses the supplied preference evidence and \mbox{current query image.} The protocol applies to methods that condition predictions on preference examples or a user representation derived from these preferences. It comprises five control tests and keeps the evaluated method, users and query samples, image-processing steps, and metric computation fixed. It changes only one condition at a time, so each quality difference measures the response to that condition.

\noindent\textbf{PCVP Controls.}
\underline{\textbf{(1)} Wrong User:} The \textit{Fixed Wrong-User Profile} replaces $p_u$, the profile of the current user $u$,
with a fixed profile from another user.
\underline{\textbf{(2)} Reversed Order:} The \textit{Reversed-Order Profile} swaps the preferred and
non-preferred image in every reference pair before profile construction.
\underline{\textbf{(3)} Mismatched Pairs:} The \textit{Mismatched-Pair Profile} breaks within-user pair correspondence while preserving
the preferred and non-preferred image sets.
\underline{\textbf{(4)} Training Mean:} The \textit{Training-User Mean Profile} replaces $p_u$ with the arithmetic mean of profiles
constructed from training users only. \underline{\textbf{(5)} Wrong Query:} The \textit{Cyclic Wrong-Query Control} conditions
the predictor on a different query from the same user. Any separate image-application
path retains the original query. Appendix~\ref{app:pcvp-details} gives the exact constructions,
preserved information, and dependencies tested for these five controls.

For a query sample $(u,q)$, let $F(p,Q_{\mathrm{cond}},Q_{\mathrm{apply}})$ denote
the complete edit under profile $p$ and conditioning image $Q_{\mathrm{cond}}$,
with $Q_{\mathrm{apply}}$ retained on any separate image-application path.
The correct and controlled outputs are defined as follows:
\begin{equation}
  \widehat I_{u,q}^{(0)}=F(p_u,Q_{u,q},Q_{u,q}),\qquad
  \widehat I_{u,q}^{(c)}=
  \begin{cases}
    F(p_u^{(c)},Q_{u,q},Q_{u,q}), & c\in\mathcal C_{\mathrm{profile}},\\
    F(p_u,Q_{u,q}^{(c)},Q_{u,q}), & c=\mathrm{query},
  \end{cases}
  \label{eq:pcvp-output}
\end{equation}
where $\mathcal C_{\mathrm{profile}}$ contains the four profile controls above.
Let $s_k(\widehat I\mid\mathcal E)$ be a metric defined so that higher values are better. The paired PCVP gain for measure $k$ and control $c$ is:
\begin{equation}
  \Delta_k^{(c)}=
  s_k\!\left(\widehat I^{(0)}\mid\mathcal E\right)
  -s_k\!\left(\widehat I^{(c)}\mid\mathcal E\right),
  \label{eq:pcvp-gain}
\end{equation}
where $\mathcal E$ is the same set of paired users and query samples for both terms.
On PPSD, $s_k$ is the comparative quality score (CQS), with
$s_k=\operatorname{CQS}_k$ for
$k\in\{\Delta E_{00},\mathrm{LPIPS},\mathrm{PSNR},\mathrm{SSIM}\}$.
We call the evaluation a PCVP pass when the lower endpoint of the paired 95\% bootstrap
interval for every $\Delta_k^{(c)}$ is above zero. When a control is repeated across
preregistered evaluation episodes, the same direction is required for the repeated estimate:
\begin{equation}
  \operatorname{PCVP\text{-}pass}
  \Longleftrightarrow
  \operatorname{LCB}_{95\%}\!\left(\Delta_k^{(c)}\right)>0
  \quad\forall k,c.
  \label{eq:pcvp-criterion}
\end{equation}
Here $\operatorname{LCB}_{95\%}$ denotes the lower endpoint of the paired bootstrap
interval. Each control passes only when all metrics satisfy this condition,
and a full PCVP pass requires all five controls to pass.

\begin{figure*}[!t]
  \centering
  \includegraphics[width=1\textwidth]{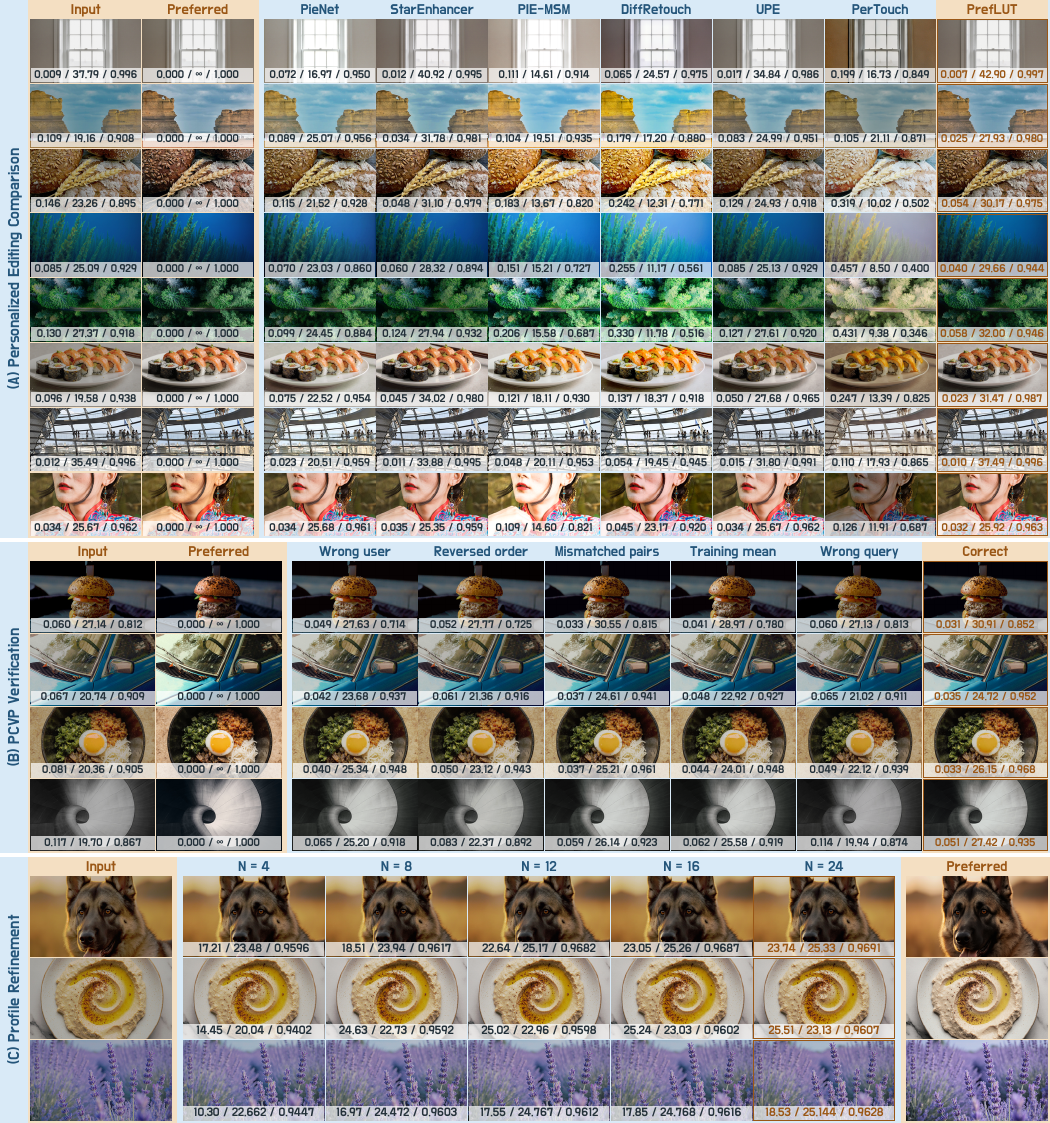}
  \caption{Qualitative results on PPSD.
  Input is the non-preferred query and Preferred its paired target.
  (A) Baseline comparisons. (B) Correct (the intended user's profile and
  current query image) versus five PCVP controls.
  (C) Profile refinement from $N=4$ to $24$ feedback pairs with frozen network weights.
  Overlays report LPIPS/PSNR/SSIM against Preferred in (A,B)
  ($\downarrow/\uparrow/\uparrow$), and per-output LPIPS-CQS/PSNR-CQS/SSIM-CQS in (C) (all $\uparrow$).
  More examples are in Appendix~\ref{app:qualitative}.}
  \label{fig:qualitative-main}
\end{figure*}

\begin{table*}[!t]
  \centering
  \caption{PCVP gains ($\Delta\mathrm{CQS}$) are correct-condition CQS minus control-condition CQS.
  Each test lists gains (left) and the number of passing metrics out of four (right).
  Bold blue marks passing metrics. A test requires 4/4, and full PCVP requires all five tests.
  A dash denotes a structurally inapplicable control. $\dagger$ marks PPSD-adapted baselines.
  Evaluation settings are in Appendix~\ref{app:pcvp-settings}.}
  \label{tab:pcvp-main}
  \small
  \definecolor{PCVPPassInk}{HTML}{335F83}
  \definecolor{PCVPWarmCell}{HTML}{FBF6EF}
  \definecolor{PCVPCoolCell}{HTML}{EEF4F9}
  \definecolor{PCVPWarmHeader}{HTML}{F3E7D8}
  \definecolor{PCVPCoolHeader}{HTML}{DDE9F3}
  \setlength{\tabcolsep}{2pt}
  \resizebox{\textwidth}{!}{%
  \begin{tabular}{l@{\hspace{6pt}}l|>{\columncolor{PCVPWarmCell}}r>{\columncolor{PCVPWarmCell}}c|>{\columncolor{PCVPCoolCell}}r>{\columncolor{PCVPCoolCell}}c|>{\columncolor{PCVPWarmCell}}r>{\columncolor{PCVPWarmCell}}c|>{\columncolor{PCVPCoolCell}}r>{\columncolor{PCVPCoolCell}}c|>{\columncolor{PCVPWarmCell}}r>{\columncolor{PCVPWarmCell}}c|>{\columncolor{PCVPCoolCell}}c}
    \toprule
  Method & Metric & \multicolumn{2}{c|}{\cellcolor{PCVPWarmHeader}Wrong user} & \multicolumn{2}{c|}{\cellcolor{PCVPCoolHeader}Reversed order} & \multicolumn{2}{c|}{\cellcolor{PCVPWarmHeader}Mismatched pairs} & \multicolumn{2}{c|}{\cellcolor{PCVPCoolHeader}Training mean} & \multicolumn{2}{c|}{\cellcolor{PCVPWarmHeader}Wrong query} & \cellcolor{PCVPCoolHeader}Full Pass \\
    \midrule\midrule
  \multirow{4}{*}{\shortstack[l]{PieNet$^{\dagger}$\\(ECCV 2020)}} & $\Delta E_{00}$ & -0.00006 &  & +0.00001 &  & +0.00068 &  & -0.00360 &  & \textcolor{PCVPPassInk}{\textbf{+0.00454}} &  &  \\
   & LPIPS & -0.07606 &  & +0.07724 &  & +0.11300 &  & -0.33478 &  & -0.35530 &  &  \\
   & PSNR & -0.10056 &  & +0.14574 &  & +0.22164 &  & -0.64456 &  & -0.21505 &  &  \\
   & SSIM & -0.00059 & \multirow{-4}{*}{0/4} & -0.00027 & \multirow{-4}{*}{0/4} & +0.00035 & \multirow{-4}{*}{0/4} & -0.00169 & \multirow{-4}{*}{0/4} & +0.00011 & \multirow{-4}{*}{1/4} & \multirow{-4}{*}{\shortstack{No\\(0/5)}} \\
    \midrule
  \multirow{4}{*}{\shortstack[l]{StarEnhancer$^{\dagger}$\\(ICCV 2021)}} & $\Delta E_{00}$ & +0.00145 &  & \textcolor{PCVPPassInk}{\textbf{+0.00334}} &  & --- &  & -0.00414 &  & \textcolor{PCVPPassInk}{\textbf{+0.02791}} &  &  \\
   & LPIPS & \textcolor{PCVPPassInk}{\textbf{+0.20574}} &  & \textcolor{PCVPPassInk}{\textbf{+0.19325}} &  & --- &  & -0.14146 &  & \textcolor{PCVPPassInk}{\textbf{+3.75054}} &  &  \\
   & PSNR & +0.14513 &  & +0.27924 &  & --- &  & -0.84752 &  & \textcolor{PCVPPassInk}{\textbf{+1.30143}} &  &  \\
   & SSIM & +0.00065 & \multirow{-4}{*}{1/4} & \textcolor{PCVPPassInk}{\textbf{+0.00123}} & \multirow{-4}{*}{3/4} & --- & \multirow{-4}{*}{---} & -0.00037 & \multirow{-4}{*}{0/4} & \textcolor{PCVPPassInk}{\textbf{+0.00982}} & \multirow{-4}{*}{\textcolor{PCVPPassInk}{\textbf{4/4}}} & \multirow{-4}{*}{\shortstack{No\\(1/4)}} \\
    \midrule
  \multirow{4}{*}{\shortstack[l]{PIE-MSM$^{\dagger}$\\(TCSVT 2024)}} & $\Delta E_{00}$ & \textcolor{PCVPPassInk}{\textbf{+0.00359}} &  & \textcolor{PCVPPassInk}{\textbf{+0.00792}} &  & \textcolor{PCVPPassInk}{\textbf{+0.03463}} &  & -0.00111 &  & \textcolor{PCVPPassInk}{\textbf{+0.01114}} &  &  \\
   & LPIPS & \textcolor{PCVPPassInk}{\textbf{+0.29696}} &  & \textcolor{PCVPPassInk}{\textbf{+0.60508}} &  & \textcolor{PCVPPassInk}{\textbf{+3.31201}} &  & +0.05089 &  & \textcolor{PCVPPassInk}{\textbf{+1.59296}} &  &  \\
   & PSNR & +0.15297 &  & \textcolor{PCVPPassInk}{\textbf{+0.51038}} &  & \textcolor{PCVPPassInk}{\textbf{+2.96831}} &  & -0.23564 &  & \textcolor{PCVPPassInk}{\textbf{+0.58833}} &  &  \\
   & SSIM & +0.00047 & \multirow{-4}{*}{2/4} & \textcolor{PCVPPassInk}{\textbf{+0.00114}} & \multirow{-4}{*}{\textcolor{PCVPPassInk}{\textbf{4/4}}} & \textcolor{PCVPPassInk}{\textbf{+0.01072}} & \multirow{-4}{*}{\textcolor{PCVPPassInk}{\textbf{4/4}}} & +0.00030 & \multirow{-4}{*}{0/4} & \textcolor{PCVPPassInk}{\textbf{+0.00617}} & \multirow{-4}{*}{\textcolor{PCVPPassInk}{\textbf{4/4}}} & \multirow{-4}{*}{\shortstack{No\\(3/5)}} \\
    \midrule
  \multirow{4}{*}{\shortstack[l]{DiffRetouch$^{\dagger}$\\(AAAI 2025)}} & $\Delta E_{00}$ & +0.00053 &  & -0.00229 &  & -0.00094 &  & +0.00014 &  & \textcolor{PCVPPassInk}{\textbf{+0.01422}} &  &  \\
   & LPIPS & +0.06096 &  & -0.22642 &  & -0.10286 &  & -0.00917 &  & \textcolor{PCVPPassInk}{\textbf{+3.05005}} &  &  \\
   & PSNR & +0.05703 &  & -0.10169 &  & -0.02648 &  & +0.03741 &  & \textcolor{PCVPPassInk}{\textbf{+1.14541}} &  &  \\
   & SSIM & +0.00032 & \multirow{-4}{*}{0/4} & -0.00267 & \multirow{-4}{*}{0/4} & -0.00091 & \multirow{-4}{*}{0/4} & +0.00005 & \multirow{-4}{*}{0/4} & \textcolor{PCVPPassInk}{\textbf{+0.02181}} & \multirow{-4}{*}{\textcolor{PCVPPassInk}{\textbf{4/4}}} & \multirow{-4}{*}{\shortstack{No\\(1/5)}} \\
    \midrule
  \multirow{4}{*}{\shortstack[l]{PerTouch$^{\dagger}$\\(AAAI 2026)}} & $\Delta E_{00}$ & +0.00227 &  & \textcolor{PCVPPassInk}{\textbf{+0.00617}} &  & -0.00106 &  & -0.00128 &  & \textcolor{PCVPPassInk}{\textbf{+0.11441}} &  &  \\
   & LPIPS & +0.14506 &  & \textcolor{PCVPPassInk}{\textbf{+0.28275}} &  & -0.06002 &  & -0.08118 &  & \textcolor{PCVPPassInk}{\textbf{+10.88617}} &  &  \\
   & PSNR & +0.10407 &  & \textcolor{PCVPPassInk}{\textbf{+0.39237}} &  & -0.06274 &  & -0.07681 &  & \textcolor{PCVPPassInk}{\textbf{+14.32032}} &  &  \\
   & SSIM & +0.00074 & \multirow{-4}{*}{0/4} & +0.00055 & \multirow{-4}{*}{3/4} & -0.00081 & \multirow{-4}{*}{0/4} & -0.00070 & \multirow{-4}{*}{0/4} & \textcolor{PCVPPassInk}{\textbf{+0.61326}} & \multirow{-4}{*}{\textcolor{PCVPPassInk}{\textbf{4/4}}} & \multirow{-4}{*}{\shortstack{No\\(1/5)}} \\
    \midrule
  \multirow{4}{*}{\shortstack[l]{User-specific\\Decoder\\(CVPR 2026)}} & $\Delta E_{00}$ & +0.00089 &  & \textcolor{PCVPPassInk}{\textbf{+0.01660}} &  & -0.00237 &  & \textcolor{PCVPPassInk}{\textbf{+0.01229}} &  & +0.00000 &  &  \\
   & LPIPS & +0.08042 &  & \textcolor{PCVPPassInk}{\textbf{+1.17893}} &  & +0.02181 &  & \textcolor{PCVPPassInk}{\textbf{+0.79596}} &  & -0.00012 &  &  \\
   & PSNR & -0.00132 &  & +0.34232 &  & -1.19466 &  & -0.01615 &  & +0.00001 &  &  \\
   & SSIM & +0.00086 & \multirow{-4}{*}{0/4} & +0.00032 & \multirow{-4}{*}{2/4} & -0.00062 & \multirow{-4}{*}{0/4} & +0.00072 & \multirow{-4}{*}{2/4} & -0.00000 & \multirow{-4}{*}{0/4} & \multirow{-4}{*}{\shortstack{No\\(0/5)}} \\
    \midrule
  \multirow{4}{*}{\shortstack[l]{UPE\\(CVPR 2026)}} & $\Delta E_{00}$ & +0.00022 &  & -0.00067 &  & -0.00073 &  & -0.00133 &  & +0.00004 &  &  \\
   & LPIPS & +0.01448 &  & -0.00233 &  & -0.00851 &  & -0.00101 &  & \textcolor{PCVPPassInk}{\textbf{+0.05344}} &  &  \\
   & PSNR & -0.01553 &  & -0.37663 &  & -0.30116 &  & -0.70954 &  & -0.22842 &  &  \\
   & SSIM & \textcolor{PCVPPassInk}{\textbf{+0.00003}} & \multirow{-4}{*}{1/4} & +0.00001 & \multirow{-4}{*}{0/4} & \textcolor{PCVPPassInk}{\textbf{+0.00001}} & \multirow{-4}{*}{1/4} & +0.00001 & \multirow{-4}{*}{0/4} & \textcolor{PCVPPassInk}{\textbf{+0.00018}} & \multirow{-4}{*}{2/4} & \multirow{-4}{*}{\shortstack{No\\(0/5)}} \\
    \midrule
  \multirow{4}{*}{\shortstack[l]{Exemplar-based\\Inference\\(CVPR 2026)}} & $\Delta E_{00}$ & +0.00024 &  & -0.00099 &  & -0.00090 &  & \textcolor{PCVPPassInk}{\textbf{+0.00090}} &  & +0.00029 &  &  \\
   & LPIPS & +0.03730 &  & -0.09915 &  & -0.10712 &  & \textcolor{PCVPPassInk}{\textbf{+0.15598}} &  & +0.02920 &  &  \\
   & PSNR & +0.02825 &  & -0.14419 &  & -0.18782 &  & \textcolor{PCVPPassInk}{\textbf{+0.19090}} &  & +0.00499 &  &  \\
   & SSIM & +0.00029 & \multirow{-4}{*}{0/4} & -0.00237 & \multirow{-4}{*}{0/4} & -0.00125 & \multirow{-4}{*}{0/4} & +0.00008 & \multirow{-4}{*}{3/4} & \textcolor{PCVPPassInk}{\textbf{+0.00162}} & \multirow{-4}{*}{1/4} & \multirow{-4}{*}{\shortstack{No\\(0/5)}} \\
    \midrule
    \midrule\midrule
  \multirow{4}{*}{\shortstack[l]{PrefLUT\\(ours)}} & $\Delta E_{00}$ & \textcolor{PCVPPassInk}{\textbf{+0.00437}} &  & \textcolor{PCVPPassInk}{\textbf{+0.00687}} &  & \textcolor{PCVPPassInk}{\textbf{+0.00723}} &  & \textcolor{PCVPPassInk}{\textbf{+0.00235}} &  & \textcolor{PCVPPassInk}{\textbf{+0.01445}} &  &  \\
   & LPIPS & \textcolor{PCVPPassInk}{\textbf{+0.44810}} &  & \textcolor{PCVPPassInk}{\textbf{+0.33821}} &  & \textcolor{PCVPPassInk}{\textbf{+0.44371}} &  & \textcolor{PCVPPassInk}{\textbf{+0.26901}} &  & \textcolor{PCVPPassInk}{\textbf{+2.26491}} &  &  \\
   & PSNR & \textcolor{PCVPPassInk}{\textbf{+0.40935}} &  & \textcolor{PCVPPassInk}{\textbf{+0.78150}} &  & \textcolor{PCVPPassInk}{\textbf{+0.90339}} &  & \textcolor{PCVPPassInk}{\textbf{+0.32984}} &  & \textcolor{PCVPPassInk}{\textbf{+0.88546}} &  &  \\
   & SSIM & \textcolor{PCVPPassInk}{\textbf{+0.00148}} & \multirow{-4}{*}{\textcolor{PCVPPassInk}{\textbf{4/4}}} & \textcolor{PCVPPassInk}{\textbf{+0.00106}} & \multirow{-4}{*}{\textcolor{PCVPPassInk}{\textbf{4/4}}} & \textcolor{PCVPPassInk}{\textbf{+0.00118}} & \multirow{-4}{*}{\textcolor{PCVPPassInk}{\textbf{4/4}}} & \textcolor{PCVPPassInk}{\textbf{+0.00078}} & \multirow{-4}{*}{\textcolor{PCVPPassInk}{\textbf{4/4}}} & \textcolor{PCVPPassInk}{\textbf{+0.00480}} & \multirow{-4}{*}{\textcolor{PCVPPassInk}{\textbf{4/4}}} & \multirow{-4}{*}{\shortstack{\textcolor{PCVPPassInk}{\textbf{Yes}}\\(5/5)}} \\
    \bottomrule
  \end{tabular}}
\end{table*}

\section{Experiments and Results}

\noindent\textbf{Implementation Details and Experimental Settings.}\quad
\underline{\textbf{(1)} Datasets:} We evaluate PrefLUT on three datasets for different purposes. We use PPSD, a recent large-scale
dataset specifically designed for learning personalized photographic styles from pairwise user
preferences, as our primary benchmark. MIT-Adobe FiveK Expert C and
PPR10K Experts A/B/C evaluate general-purpose (non-personalized) aesthetic enhancement
(Appendix~\ref{app:transfer-comparisons}).
\underline{\textbf{(2)} Metrics:} On PPSD, personalized enhancement is evaluated using CQS (Appendix~\ref{app:metric-statistics}),
computed separately from $\Delta E_{00}$, LPIPS, PSNR, and SSIM,
combining image fidelity with color movement toward user preferences to assess
both preservation and preference alignment~\citep{kim2026ppsd}.
Full dataset and evaluation protocols are provided in
Appendices~\ref{app:datasets} and~\ref{app:pcvp}.
\underline{\textbf{(3)} Experimental Settings:} PPSD evaluation uses 50 users excluded from training, each with 16 ordered reference
pairs and 16 query pairs with disjoint image and scene identities.
The standard PrefLUT setting uses a 256-dimensional Reusable User Profile and a $17^3$ LUT
(Appendix~\ref{app:preflut-settings}).
All our experiments are conducted on a single NVIDIA RTX~5090 GPU.
\underline{\textbf{(4)} Baselines:} We select personalized enhancement methods and controllable
retouching methods adaptable to user preferences, including PieNet~\citep{kim2020pienet}, StarEnhancer~\citep{song2021starenhancer},
PIE-MSM~\citep{kosugi2023masked}, DiffRetouch~\citep{duan2024diffretouch},
PerTouch~\citep{chang2026pertouch}, and PPSD's User-specific Decoder,
User Preference Embedding (UPE), and Exemplar-based Inference~\citep{kim2026ppsd}.
Baseline reproduction and evaluation
settings are detailed in Appendix~\ref{app:training}.

\noindent\textbf{PPSD Personalized Editing.}
PrefLUT balances perceptual quality, per-user storage, and editing latency
(Table~\ref{tab:adapted-personalized} and Appendix~\ref{app:efficiency-settings}), with the highest CQS under $\Delta E_{00}$, LPIPS, and SSIM
among the listed methods. On edited non-preferred queries, PrefLUT also leads
the compared methods in direct fidelity to Preferred under $\Delta E_{00}$, LPIPS, PSNR, and SSIM.
For qualitative evaluation, Figure~\ref{fig:qualitative-main}(A) compares outputs
across baselines and illustrates PrefLUT's robustness to varying preferences.
For example, users prefer both brighter and darker appearances, yet PIE-MSM and
DiffRetouch tend to brighten these examples, while PrefLUT often follows
the preferred brightness and color more closely
(additional examples in Appendix~\ref{app:qualitative}).

\noindent\textbf{PCVP Evaluation.}\quad
PrefLUT passes all five PPSD controls across
all four CQS measures (Table~\ref{tab:pcvp-main}). None of the eight baseline
adaptations passes all five. Relative to the correct condition, all five controls
significantly reduce CQS, indicating that PrefLUT uses both the supplied user
preferences and the query image (Appendix~\ref{app:baseline-pcvp-complete}).
In Figure~\ref{fig:qualitative-main}(B), changing the profile or query condition
shifts brightness and color away from the preferred appearance, while correct
conditioning more closely preserves the intended look.

\begin{figure}[!t]
  \centering
  \includegraphics[width=\textwidth]{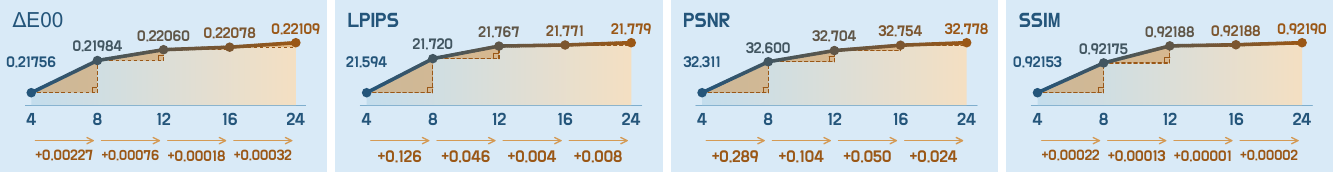}
  \caption{Profile refinement on PPSD. Points show CQS (all $\uparrow$)
  at $N=4,8,12,16,24$ accumulated feedback pairs. Orange labels give CQS gains
  over the preceding stage.}
  \label{fig:profile-refinement-main}
\end{figure}

\noindent\textbf{Profile Refinement.}\quad
With weights and unseen queries fixed, additional feedback improves editing through
profile updates alone (Figure~\ref{fig:profile-refinement-main},
Figure~\ref{fig:qualitative-main}(C)). The examples progressively approach
the preferred brightness and color as feedback accumulates from 4 to 24 pairs.
Correct new feedback outperforms reversed new feedback on identical images,
indicating that the preference direction contributes to the improvement
(Appendix~\ref{app:profile-refinement}).

\noindent\textbf{Ablation Studies.}\quad
Table~\ref{tab:architecture-ablation-main} compares each ablation with its matched
full model under equal training budgets. Removing the Explicit Feature Difference
from the Ordered Preference Encoder reduces all four CQS measures and the quality
advantage of correct over wrong-user profiles. Replacing the Set Transformer in
the Preference Set Aggregator with masked mean pooling reduces CQS and nearly
eliminates this advantage, supporting learned aggregation of preference pairs.
Removing the Query Feature from the Query-Conditioned LUT Predictor reduces
$\Delta E_{00}$, LPIPS, and SSIM CQS and weakens user-specific editing, despite higher
PSNR CQS. The Wrong-User Contrast Loss further improves the wrong-user
PCVP gain across both training seeds.
Paired confidence intervals and further ablations are in
Appendices~\ref{app:core-ablations} and~\ref{app:profile-width}.

\noindent\textbf{Extended Experiments.}\quad
The profile interface supports general-purpose, non-personalized enhancement
on FiveK and PPR10K using configurations with an additional spatial residual
(Appendices~\ref{app:task-settings} and~\ref{app:transfer-comparisons}).
On FiveK Expert C, PrefLUT achieves 24.572\,dB PSNR,
0.917 SSIM, 7.833 $\Delta E_{76}$, and 0.067 LPIPS, averaged over eight training seeds.
On PPR10K, PrefLUT shares one model across Experts A/B/C and achieves
25.105\,dB PSNR, 7.822 $\Delta E_{76}$, 28.368\,dB PSNR-HC, and
5.069 $\Delta E_{76}$-HC, averaged over three experts and two training seeds
on \mbox{2,286 validation images per expert.}
\begin{table}[!t]
  \centering
  \caption{Core ablations on PPSD. Differences in CQS and Fixed Wrong-User Profile
  PCVP gain are reported as variant minus its matched full model.
  In row order, we remove the Explicit Feature Difference $a_i^+-a_i^-$,
  replace the Set Transformer with masked mean pooling in the Preference Set Aggregator,
  remove the Query Feature $c_q$, and remove the Wrong-User Contrast Loss
  $\mathcal L_{\mathrm{user}}$. All ablations support the contribution of the corresponding components.}
  \label{tab:architecture-ablation-main}
  \small
  \definecolor{AblationQualityA}{HTML}{FBF6EF}
  \definecolor{AblationQualityB}{HTML}{F3E7D8}
  \definecolor{AblationQualityHeader}{HTML}{F0DFC9}
  \definecolor{AblationGainA}{HTML}{F3F7FB}
  \definecolor{AblationGainB}{HTML}{E3EDF6}
  \definecolor{AblationGainHeader}{HTML}{DDE9F3}
  \setlength{\tabcolsep}{3pt}
  \renewcommand{\arraystretch}{1.08}
  \resizebox{\columnwidth}{!}{%
  \begin{tabular}{@{}l|*{2}{>{\columncolor{AblationQualityA}}r>{\columncolor{AblationQualityB}}r}|*{2}{>{\columncolor{AblationGainA}}r>{\columncolor{AblationGainB}}r}@{}}
    \toprule
   & \multicolumn{4}{c|}{\cellcolor{AblationQualityHeader}CQS difference} & \multicolumn{4}{c}{\cellcolor{AblationGainHeader}Wrong-user PCVP gain difference} \\
  Variant & $\Delta E_{00}$ & LPIPS & PSNR & SSIM & $\Delta E_{00}$ & LPIPS & PSNR & SSIM \\
    \midrule
  $-$ Explicit Feature Difference & $-0.001062$ & $-0.200$ & $-0.214$ & $-0.000553$ & $-0.003074$ & $-0.32296$ & $-0.28613$ & $-0.001116$ \\
  $-$ Set Transformer & $-0.019845$ & $-1.484$ & $-2.250$ & $-0.002619$ & $-0.004238$ & $-0.42888$ & $-0.39226$ & $-0.001401$ \\
  $-$ Query Feature & $-0.003771$ & $-1.438$ & $+1.482$ & $-0.003514$ & $-0.003290$ & $-0.33886$ & $-0.34294$ & $-0.001110$ \\
  $-$ Wrong-User Contrast Loss & $-0.00000679$ & $-0.003$ & $+0.003$ & $-0.00000439$ & $-0.00003627$ & $-0.00487$ & $-0.00223$ & $-0.00001351$ \\
    \bottomrule
  \end{tabular}}
\end{table}

\section{Conclusion}

We presented PrefLUT, which encodes ordered preferences into a reusable, refinable
user profile and predicts query-conditioned 3D LUTs without per-user fine-tuning.
We also introduced PCVP, an evaluation protocol that verifies dependence on user
preferences and the query image through five controlled tests. This distinguishes
user-specific preference conditioning from quality gains produced by applying
shared editing rules across users. Additional feedback
improves unseen-query editing with frozen network weights, while each quantized
profile requires only 260 bytes and supports repeated editing without re-encoding
reference images. PrefLUT passes all five PCVP tests and, among the compared personalized image enhancement methods on PPSD, leads all four direct fidelity metrics and CQS under
$\Delta E_{00}$, LPIPS, and SSIM, with editing at 1.365\,ms/image.
Experiments on FiveK and PPR10K further demonstrate general-purpose enhancement
across expert targets.

\subsection*{AI Use Statement}
Generative AI tools assisted manuscript writing and the development of code for
figure layout. Qualitative figures show images from the released datasets and
actual outputs of the evaluated methods. These dataset images and method outputs
were cropped and resized for display.

\subsection*{Ethics Statement}
This work uses released datasets of pairwise preferences and retouched images
and conducts no new interaction with human participants. We follow the dataset
licenses and do not redistribute raw user data. Dataset splits and evaluation
procedures are documented in the appendix. This research adheres to the ICLR Code
of Ethics.

\subsection*{Reproducibility Statement}
The main paper specifies the PrefLUT architecture, training objectives, and evaluation protocol.
The appendix details data preprocessing, dataset splits, experimental configurations, control
conditions, statistical procedures, and hardware. The supplementary material includes anonymous
source code, training and evaluation scripts, configuration files, PPSD split definitions,
exact reference/query assignments, and the PrefLUT model weights. We also provide
complete per-query and per-user evaluation CSVs, aggregate results, and selected qualitative
examples with preference references. The accompanying documentation maps the implementation
to the paper and provides instructions for training, profile construction, inference, and
metric computation.

\bibliographystyle{iclr2027_conference}
\bibliography{references}

\clearpage
\appendix
\begin{center}
  {\fontsize{24}{28}\selectfont\bfseries Appendix\par}
\end{center}
\vspace{6pt}
\definecolor{AppBlueA}{HTML}{EDF5FB}
\definecolor{AppBlueB}{HTML}{DCEBF6}
\definecolor{AppBlueHeader}{HTML}{C7DFEF}
\definecolor{AppOrangeA}{HTML}{FBF6EF}
\definecolor{AppOrangeB}{HTML}{F3E7D8}
\definecolor{AppOrangeHeader}{HTML}{F0DFC9}
\definecolor{AppPassInk}{HTML}{335F83}
\newcommand{\appgain}[1]{\textcolor{AppPassInk}{#1}}
\newlength{\AppMetricWidth}

\makeatletter
\@addtoreset{table}{section}
\@addtoreset{figure}{section}
\makeatother
\renewcommand{\thetable}{\Alph{section}\arabic{table}}
\renewcommand{\thefigure}{\Alph{section}\arabic{figure}}
\setcounter{table}{0}
\setcounter{figure}{0}
\renewcommand{\topfraction}{0.95}
\renewcommand{\bottomfraction}{0.95}
\renewcommand{\textfraction}{0.05}
\renewcommand{\floatpagefraction}{0.85}
\setcounter{topnumber}{5}
\setcounter{totalnumber}{8}
\section{Method Details}
\label{app:protocol}

\subsection{Profile Construction and Query-Time Editing}
\label{app:method-data-flow}

Algorithm A1 summarizes Profile Construction and Query-Time Editing, including
the inputs and outputs of each module. Appendix~\ref{app:batching} explains how
reference sets with different numbers of pairs are processed in one batch.

\begin{figure*}[!htbp]
  \centering
  \setlength{\fboxsep}{6pt}
  \fbox{\begin{minipage}{\dimexpr\textwidth-2\fboxsep-2\fboxrule\relax}
  \textbf{Algorithm A1: profile construction and query-time editing}\\[3pt]
  \textbf{Profile-construction input:} reference set
  $S_u=\{(I_i^+,I_i^-)\}_{i=1}^{N_u}$. When users have different numbers of reference pairs,
  shorter sets are padded to a common length. The mask $m_{u,i}$ marks valid pairs.\\
  \textbf{1.} Encode both images of every valid pair with the shared Reference Image Encoder,
  producing $a_i^+$ and $a_i^-$.\\
  \textbf{2.} Compute $t_i=\phi([a_i^+,a_i^-,a_i^+-a_i^-])$ with the Ordered
  Preference Encoder.\\
  \textbf{3.} Place the valid ordered preference tokens in deterministic order and jointly update
  them with the learned pooling tokens in the Preference Set Aggregator. Use the valid-pair mask to exclude
  padding and do not add positional embeddings.\\
  \textbf{4.} Average and layer-normalize the updated learned pooling tokens to obtain the
  aggregated user feature $r_u$. Apply Profile Projection
  $p_u=B_{\mathrm{down}}(r_u)$ and store the resulting Reusable User Profile $p_u$.
  The reference images are no longer needed at query time.\\[3pt]
  \textbf{Query-time input:} Reusable User Profile $p_u$ and query image $I_q$.\\
  \textbf{5.} Apply Profile Expansion $\bar p_u=B_{\mathrm{up}}(p_u)$. Resize $I_q$
  to the query thumbnail $I_q^\downarrow$ and use the Query Image Encoder to obtain the
  query feature $c_q$.\\
  \textbf{6.} Use the Query-Conditioned LUT Predictor to predict the LUT latent vector
  $z_{u,q}$ and edit strength $g_{u,q}$ from $[\bar p_u,c_q]$.\\
  \textbf{7.} Use the Identity-Residual LUT Decoder to convert $z_{u,q}$ into the
  decoded LUT $D(z_{u,q})$. The Edit-Strength Controller scales the decoded LUT
  residual from the identity LUT by $g_{u,q}$ to obtain the final LUT $\widehat L_{u,q}$.\\
  \textbf{8.} Apply $\widehat L_{u,q}$ to the full-resolution query image by trilinear
  interpolation to obtain $\widehat I_{u,q}$.
  \end{minipage}}
  \par\smallskip
  \begin{minipage}{0.96\textwidth}
  \small\textbf{Profile refinement and reuse.} Add new preference pairs to the retained
  reference set and repeat Steps 1 to 4 with network weights frozen. Steps 5 to 8
  reuse the updated profile for later queries without processing references or
  performing per-user optimization.
  \end{minipage}
\end{figure*}

\subsection{Batching and Valid-Pair Masking}
\label{app:batching}

In a batch, the valid-pair mask $m_{u,i}$ identifies whether ordered preference
image pair $i$ belongs to user $u$'s reference set. Padding makes reference sets
equally long, and the mask excludes padded tokens from attention and profile
aggregation. Learned pooling tokens are always valid. Equation~\ref{eq:setaggregation}
therefore uses only valid ordered preference tokens. The Preference Set Aggregator
uses a fixed token order and no positional embeddings.

Valid tokens are sorted in ascending order by their first eight feature coordinates,
comparing each coordinate in turn. Tokens with identical sorting coordinates keep
their relative order, and padded tokens are placed last. This fixed order is based
on token features rather than the time of user feedback. The aggregated user feature
$r_u$ is computed from the updated learned pooling tokens. Sorting reduces
input-order effects in finite-precision attention. Permutation invariance comes
from the set architecture without positional embeddings, rather than from sorting.

\subsection{Target LUT Fitting}
\label{app:target-lut-fitting}

For each ordered preference image pair identified as pixel-aligned by the data
protocol, a target LUT $L^*$ maps the non-preferred image to the preferred image.
The training subset contains 2,683 eligible D1 pairs and 3,028 eligible D2 pairs,
giving 5,711 target LUTs. Eligibility requires a D1 or D2 pair with matching image
dimensions. Each $17^3$ LUT is initialized as the identity LUT and fitted on
128-pixel image crops using Adam for 30 steps at learning rate 0.03.
The objective combines mean absolute image reconstruction error with smoothness,
monotonicity, and identity penalties weighted by 1, 0.01, 0.01, and 0.001,
respectively. After each update, LUT values are clipped to $[0,1]$. The fitted LUTs supervise
Identity-Residual LUT Decoder pretraining. Query-Time Editing uses the trained
decoder without fitting additional target LUTs.

\subsection{Identity-Residual LUT Decoder Pretraining}
\label{app:decoder-pretraining}

The fitted target LUTs supervise an autoencoder whose Identity-Residual LUT Decoder
follows Equation~\ref{eq:decoder}. Let $E_L$ be the LUT encoder, let
$\overline L=L_{\mathrm{id}}+\beta\tanh(\Psi(E_L(L^*)))$ be the unclipped decoder
output, let $\widetilde L=\operatorname{clip}_{[0,1]}(\overline L)=D(E_L(L^*))$, and let
$I$ be the non-preferred image used to fit $L^*$.
$E_L$ and the Identity-Residual LUT Decoder are trained jointly with LUT reconstruction, image application,
smoothness, monotonicity, and a LUT range penalty:
\begin{equation}
  \begin{aligned}
  \mathcal L_{\mathrm{AE}}={}&
  \eta_{\mathrm{rec}}\|\widetilde L-L^*\|_1
  +\eta_{\mathrm{app}}\|\mathcal T(\widetilde L,I)-\mathcal T(L^*,I)\|_1\\
  &+\eta_{\mathrm{sm}}\mathcal R_{\mathrm{sm}}(\widetilde L)
  +\eta_{\mathrm{mono}}\mathcal R_{\mathrm{mono}}(\widetilde L)
  +\eta_{\mathrm{range}}\mathcal R_{\mathrm{range}}(\overline L).
  \end{aligned}
  \label{eq:aeloss}
\end{equation}
$\mathcal R_{\mathrm{sm}}$ sums the mean squared differences between adjacent LUT
entries along the three input-color axes. $\mathcal R_{\mathrm{mono}}$ penalizes negative
adjacent differences, and $\mathcal R_{\mathrm{range}}$ penalizes values of $\overline L$
that fall below zero or above one.
During personalized training, the Query-Conditioned LUT Predictor outputs
$z_{u,q}$, which the frozen Identity-Residual LUT Decoder converts into the decoded
LUT. Equation~\ref{eq:loss} combines image-based supervision with edit-strength
supervision and LUT regularization, without requiring fitted target LUTs for the
preference pairs.

\subsection{Global Color-Statistics Descriptor}
\label{app:color-statistics}

For an RGB image $I$, the global color-statistics descriptor contains the means and
standard deviations of the RGB channels, luminance, and saturation:
\begin{equation}
\begin{aligned}
\chi(I)={}&[\mu_R(I),\mu_G(I),\mu_B(I),\sigma_R(I),\sigma_G(I),\sigma_B(I),\\
&\mu_Y(I),\sigma_Y(I),\mu_S(I),\sigma_S(I)].
\end{aligned}
\label{eq:descriptor-components}
\end{equation}
Here $\mu$ and $\sigma$ are the mean and population standard deviation over image pixels.
The luminance and saturation at each pixel are computed as:
\begin{equation}
Y=0.2126R+0.7152G+0.0722B,\qquad
S=\max(R,G,B)-\min(R,G,B).
\label{eq:descriptor-channels}
\end{equation}
The distance $d_\chi(A,B)$ averages the absolute differences between the ten
descriptor components. RGB, luminance, and saturation lie in $[0,1]$, so their
means and standard deviations are bounded in the same intensity units. Equal
coefficients assign the same cost to an equal absolute change in any statistic,
without variance normalization or learned component weights.
Global statistics reduce sensitivity to spatial misalignment
but do not impose pixelwise correspondence. Pixel-aligned pairs also receive reconstruction supervision (Appendix~\ref{app:auxiliary-losses}).

\subsection{Auxiliary Training Losses}
\label{app:auxiliary-losses}

The following losses supplement the color-distance loss, ranking loss, and
Wrong-User Contrast Loss in Equation~\ref{eq:loss}. Let $\mathcal B$ index all
query pairs in the current minibatch, with $(u,j)$ identifying user $u$ and query
pair $j$. Training uses four query pairs per user. The outputs are
$Y_{u,j}^\pm=\widehat I(p_u,Q_{u,j}^\pm)$. The color-distance, ranking, and
Wrong-User Contrast Loss terms are averaged over $\mathcal B$. Image and LUT
$\ell_1$ terms below average over their pixels or grid entries and color channels.

\noindent\textbf{Aligned-Pair Reconstruction Loss.}\quad
Let $\mathcal A\subseteq\mathcal B$ index the query pairs treated as pixel-aligned
by the data protocol. The aligned-pair reconstruction loss is:
\begin{equation}
\mathcal L_{\mathrm{aligned}}=|\mathcal A|^{-1}
\sum_{(u,j)\in\mathcal A}\|Y_{u,j}^- - Q_{u,j}^+\|_1.
\label{eq:aligned-reconstruction}
\end{equation}
The reconstruction loss is zero when the minibatch contains no aligned query pairs.

\noindent\textbf{Preservation Loss.}\quad
The preservation loss keeps the edited preferred query close to its input:
\begin{equation}
\mathcal L_{\mathrm{preserve}}=\frac{1}{|\mathcal B|}
\sum_{(u,j)\in\mathcal B}\|Y_{u,j}^+-Q_{u,j}^+\|_1.
\label{eq:preserve-app}
\end{equation}

\noindent\textbf{Edit-Strength Loss.}\quad
Let $g_{u,j}^\pm$ and $\widehat L_{u,j}^\pm$ be the edit strengths and final LUTs predicted
for $Q_{u,j}^\pm$. Binary supervision assigns strength targets of one to
non-preferred queries and zero to preferred queries:
\begin{equation}
\mathcal L_{\mathrm{strength}}=\frac{1}{2|\mathcal B|}
\sum_{(u,j)\in\mathcal B}\big[
\operatorname{BCE}(g_{u,j}^-,1)+\operatorname{BCE}(g_{u,j}^+,0)\big],
\label{eq:strength-app}
\end{equation}
where BCE denotes binary cross-entropy.

\noindent\textbf{LUT-Deviation Loss.}\quad
The LUT-deviation loss penalizes differences between each final LUT and the identity LUT:
\begin{equation}
\mathcal L_{\mathrm{LUT}}=\frac{1}{2|\mathcal B|}
\sum_{(u,j)\in\mathcal B}\big(
\|\widehat L_{u,j}^{-}-L_{\mathrm{id}}\|_1+
\|\widehat L_{u,j}^+-L_{\mathrm{id}}\|_1\big).
\label{eq:lut-deviation-app}
\end{equation}

\subsection{Wrong-User Profile Selection}
\label{app:wrong-user-selection}

The Wrong-User Contrast Loss uses minibatches of distinct users. A fixed nonzero
cyclic shift within each minibatch pairs each user $u$ with another user $u'\neq u$.
The other user's profile $p_{u'}$ replaces $p_u$, while the query $Q_u^-$ and preferred
target $Q_u^+$ remain unchanged. The resulting output $\widetilde Y_{u\leftarrow u'}^-$ defines
$\widetilde d^+=d_\chi(\widetilde Y_{u\leftarrow u'}^-,Q_u^+)$ in
Equation~\ref{eq:usercontrast}. Thus, the two outputs are compared on the same
query and preferred target, with only the user profile changed.

\section{Datasets and Evaluation Protocols}
\label{app:datasets}

\subsection{PPSD}
\label{app:ppsd-evaluation-protocol}

\paragraph{Preference Supervision.}
Rating-based datasets such as Flickr-AES capture overall aesthetic judgments,
which depend on image content and aesthetic attributes~\citep{ren2017personalized}.
A single score can mix color preference with content or composition, and its
numerical scale depends on the rater. Guiding enhancement with a learned scorer
can carry these biases into editing. Existing rating labels and personalized
predictors also represent some users' preferences better than
others~\citep{goree2023correct}. PrefLUT uses PPSD's pairwise choices to specify
the preferred direction without assigning an absolute aesthetic score or
optimizing a separate scorer.

\paragraph{Data and Split.}
The processed PPSD release contains 34,699 pairwise comparisons from 521 users
after filtering. The public release provides the dataset but not the official
user split or evaluation-episode generator~\citep{kim2026ppsd}. We therefore define
our own deterministic user split and evaluation episodes. The split assigns
471 users to training and 50 to validation. Validation users are excluded from
training and construction of the Training-User Mean Profile. Checkpoint selection
is described in Appendix~\ref{app:preflut-settings}.

An evaluation episode specifies the reference and query pairs sampled for every
user. The main quality and PCVP episodes use $N=16$ reference pairs and $M=16$
query pairs per user. The two sets share no pair, scene, or image identities.
An episode seed fixes the sampling and ordering of these pairs. The adapted
baselines follow the same protocol, with implementation details in
Appendix~\ref{app:baseline-reproduction}.

\paragraph{Resolution.}
For PrefLUT, the main quality comparison and Fixed Wrong-User Profile test encode reference
images at 256 pixels and apply the predicted $17^3$ LUTs to 512-pixel query images.
Outputs and targets are resized to 256 pixels for metric computation. The
Reversed-Order Profile and Mismatched-Pair Profile tests apply LUTs at 256 pixels.
The Training-User Mean Profile and Cyclic Wrong-Query Control tests apply LUTs at
the original image resolution and compute metrics at 256 pixels. Within each test,
the correct condition and the control use identical image-processing steps.
Baseline resolutions and repeated-episode settings are specified in
Appendix~\ref{app:pcvp-settings}.

\paragraph{CQS Interpretation.}
CQS combines the Base Fidelity Score with the Comparative Margin Ratio, following
PPSD~\citep{kim2026ppsd}. Its full definition and aggregation are given in
Appendix~\ref{app:metric-statistics}. On PPSD, PCVP uses
the five PCVP controls defined in Section~\ref{sec:pcvp} and Appendix~\ref{app:pcvp-details}.

\subsection{MIT-Adobe FiveK}
\label{app:fivek-dataset}

FiveK Expert C uses 4,500 training images and 500 test images. Model development
uses 4,050 images for training and 450 for validation within the training set. The selected architecture and
training duration are fixed before training on all 4,500 images. One Expert-C
profile is constructed from 16 fixed training pairs, with the source as
non-preferred and the expert-retouched image as preferred. Test images are excluded
from profile construction and model selection. Training uses paired Lightroom
sRGB images, and evaluation uses the LPTN 480p test pairs while preserving image aspect ratios.
Metrics are computed on 8-bit outputs. The reported SSIM uses Wang-2009 automatic
downsampling, and LPIPS uses AlexNet. Results average all eight predefined training
seeds, 2029 to 2036.

\subsection{PPR10K}
\label{app:ppr10k-dataset}

PPR10K is evaluated on the same released 2,286-image validation split for Experts
A, B, and C~\citep{liang2021ppr10k}. PrefLUT shares one model across all three
experts, with one 16-pair reference set constructed for each expert from training
images. Appendix~\ref{app:task-settings} details the architecture and training.

Evaluation uses the released 360-pixel sources, 8-bit outputs, and the same
implementations of PSNR, $\Delta E_{76}$, and their human-centered (HC) variants
for all experts. The HC variants use the released human mask. RGB and Lab errors
retain unit weight inside the mask and are multiplied by 0.5 outside it before
PSNR-HC and $\Delta E_{76}$-HC are computed. Results average training seeds
2034 and 2036.

\subsection{Metrics and Statistical Testing}
\label{app:metric-statistics}

\paragraph{Comparative Quality Score.}
Following Section 4.5 of PPSD~\citep{kim2026ppsd}, CQS combines the Base Fidelity
Score (BFS) and Comparative Margin Ratio (CMR). It is computed separately for
$\Delta E_{00}$~\citep{sharma2005ciede}, LPIPS~\citep{zhang2018lpips}, PSNR,
and SSIM~\citep{wang2004ssim}. For query pair $i$, let $I_p^{(i)}$ and
$I_n^{(i)}$ denote the preferred and non-preferred images. The evaluated method
edits both under the same user condition, producing $\hat I_p^{(i)}$ and
$\hat I_n^{(i)}$. For metric $m_k$, the four comparisons are:
\begin{equation}
\begin{aligned}
d_{pp,k}^{(i)}&=m_k(\hat I_p^{(i)},I_p^{(i)}), &
d_{pn,k}^{(i)}&=m_k(\hat I_p^{(i)},I_n^{(i)}),\\
d_{np,k}^{(i)}&=m_k(\hat I_n^{(i)},I_p^{(i)}), &
d_{nn,k}^{(i)}&=m_k(\hat I_n^{(i)},I_n^{(i)}).
\end{aligned}
\label{eq:cqs-four-way}
\end{equation}
The first subscript identifies the input and the second identifies the target.
Using both inputs evaluates adjustment of non-preferred images and preservation
of preferred images. For the aggregate results reported here, the raw scores
are averaged across the preferred and non-preferred inputs and all $Q$ query pairs before the
nonlinear CQS calculation:
\begin{equation}
\bar d_{p,k}=\frac{1}{2Q}\sum_{i=1}^{Q}
\left(d_{pp,k}^{(i)}+d_{np,k}^{(i)}\right),\qquad
\bar d_{n,k}=\frac{1}{2Q}\sum_{i=1}^{Q}
\left(d_{pn,k}^{(i)}+d_{nn,k}^{(i)}\right).
\label{eq:cqs-aggregation}
\end{equation}
The standard episode has $Q=800$, with 16 query pairs per user, so users receive equal
weight. For lower-is-better metrics, $k\in\{\Delta E_{00},\mathrm{LPIPS}\}$:
\begin{equation}
\mathrm{BFS}_k=\frac{1}{\sqrt{\bar d_{p,k}\bar d_{n,k}}},\qquad
\mathrm{CMR}_k=\frac{\bar d_{n,k}-\bar d_{p,k}}{\bar d_{n,k}+\bar d_{p,k}}.
\label{eq:cqs-distance}
\end{equation}
For higher-is-better metrics, $k\in\{\mathrm{PSNR},\mathrm{SSIM}\}$:
\begin{equation}
\mathrm{BFS}_k=\sqrt{\bar d_{p,k}\bar d_{n,k}},\qquad
\mathrm{CMR}_k=\frac{\bar d_{p,k}-\bar d_{n,k}}{\bar d_{p,k}+\bar d_{n,k}}.
\label{eq:cqs-similarity}
\end{equation}
The final score is:
\begin{equation}
\mathrm{CQS}_k=\mathrm{BFS}_k\left(1+\mathrm{CMR}_k\right).
\label{eq:cqs-final}
\end{equation}
BFS measures fidelity to both targets, while positive CMR indicates closer
agreement with the preferred target. All four CQS measures are higher-is-better.
The four CQS values are reported separately. The implementation
adds $10^{-12}$ to the CMR denominator and floors the reciprocal BFS denominator
at $10^{-12}$ for numerical stability. Bootstrap samples repeat the aggregation
and CQS calculation rather than averaging per-query CQS values. Per-output CQS
in qualitative figures instead uses that output's two target comparisons.
Direct Fidelity to Preferred in Table~\ref{tab:adapted-personalized} reports only
$Q^{-1}\sum_i d_{np,k}^{(i)}$, which measures edited non-preferred inputs against
preferred targets and is distinct from both $\bar d_{p,k}$ and CQS.

\paragraph{Color-Difference Notation.}
The CIE 1976 color difference is the Euclidean distance in CIELAB,
$\Delta E_{76}=\sqrt{(\Delta L^*)^2+(\Delta a^*)^2+(\Delta b^*)^2}$,
also denoted $\Delta E^*_{ab}$~\citep{cie2020vocabulary}.
The published results in Tables~\ref{tab:fivek-personalized} and~\ref{tab:ppr10k-preflut}
retain the notation $\Delta E_{ab}$, while our evaluations use $\Delta E_{76}$.
Both refer to CIELAB Euclidean color difference. Values still depend on image
processing and averaging within each protocol. This metric differs from PPSD's
CIEDE2000 $\Delta E_{00}$~\citep{sharma2005ciede}.

\paragraph{Other Metrics and Statistical Testing.}
FiveK reports PSNR, SSIM, $\Delta E_{76}$, and Alex-LPIPS, with means and sample
standard deviations over eight training seeds. PPR10K reports PSNR and $\Delta E_{76}$ together with
their human-centered variants for configurations that use expert profiles. Each
expert's result is averaged over two training seeds, followed by an equal average
across the three experts.

PPSD uses paired user-bootstrap sampling. Each resample selects 50 users with
replacement and includes all queries for each selected user under both conditions.
When several evaluation episodes are combined, measurements are grouped by user.
Control-specific resample counts are given in Appendix~\ref{app:pcvp-settings}.
For PPSD paired comparisons, a gain is statistically positive when the lower
endpoint of its paired 95\% user-bootstrap interval is above zero.

\section{Experimental Settings}
\label{app:training}

Training uses BF16 for StarEnhancer, PIE-MSM, DiffRetouch, and PerTouch.
The diffusion baselines' quality and PCVP evaluation also use BF16.
PrefLUT refinement and the PrefLUT FiveK/PPR10K configurations use FP32. Efficiency measurement settings are
specified in Appendix~\ref{app:efficiency-settings}. Compact PrefLUT deployment uses FP32 weights,
int8 profiles with one FP32 scale, and uint8 LUTs.

\subsection{PrefLUT Settings}
\label{app:preflut-settings}

\noindent\textbf{Architecture.}
The standard PrefLUT configuration uses $N=16$ ordered preference image pairs.
The Reference Image Encoder and Query Image Encoder each use four stride-2
$3\times3$ convolutional blocks with 32, 64, 128, and 256 output channels,
GroupNorm, and SiLU. Global average pooling and a $256\rightarrow256$ linear
projection produce each image feature. The Ordered Preference Encoder uses
a $768\rightarrow512\rightarrow256$ projection to form each ordered preference token. The Preference Set Aggregator
has $K=4$ learned pooling tokens, $L=4$ Transformer blocks, eight attention heads,
and no positional embeddings. The aggregated user feature and Reusable User Profile
have widths $d_f=d_p=256$. Profile Projection and Profile Expansion are identity
mappings in this standard configuration. When $d_p\neq d_f$, they are learned
linear maps without biases.
The expanded user feature and query feature feed the LUT latent vector head
($512\rightarrow512\rightarrow256$) and edit-strength head
($512\rightarrow256\rightarrow1$). The Identity-Residual LUT Decoder linearly maps
the $d_z=256$ latent vector to a $128\times4\times4\times4$ feature volume.
Three $3^3$ convolutions map channels $128\rightarrow64\rightarrow32\rightarrow3$,
with SiLU after the first two. Trilinear interpolation with aligned corners
resizes the output to $17^3$, followed by $\tanh$, residual scaling by
$\beta=0.5$, identity addition, and clipping. Descriptor computation and token
ordering follow Appendices~\ref{app:color-statistics} and~\ref{app:batching}.

\noindent\textbf{Pretraining.}
The LUT autoencoder is pretrained for 50 epochs on 5,711 fitted training LUTs
(Appendix~\ref{app:target-lut-fitting}), with 64-pixel images for its image-application
loss. Its loss weights
$(\eta_{\rm rec},\eta_{\rm app},\eta_{\rm sm},\eta_{\rm mono},\eta_{\rm range})$
are $(1,1,10^{-3},10^{-2},10^{-1})$.
An image-pair-to-LUT model is then pretrained for 30 epochs on the same aligned
training pairs and fitted LUTs, using 128-pixel images and the frozen autoencoder.
Given a preferred/non-preferred pair, it predicts a LUT latent vector whose target
is $E_L(L^*)$. The objective combines latent $\ell_1$ error, decoded-LUT $\ell_1$
error, image-application $\ell_1$ error, an order-direction hinge loss, and a confidence
loss, with weights $(1,1,1,0.1,0.05)$. The image target is the non-preferred image
transformed by $L^*$. The direction term uses margin 0.1 to favor the correctly ordered pair
over its reversal in latent distance to $E_L(L^*)$. Binary cross-entropy supervises
the confidence head with the fitted LUT's relative image-$\ell_1$ improvement over
identity, clipped to $[0,1]$.
Both pretraining stages use batch size 32, AdamW at $2\times10^{-4}$, cosine decay,
weight decay $10^{-4}$, gradient clipping at 1.0, and seed 2026. Checkpoint selection
minimizes total validation loss on 413 fitted validation LUTs and their paired
images, selecting epochs 50 and 30, respectively.

The pretrained pair encoder initializes the Ordered Preference Encoder,
including the Reference Image Encoder and pair projection. Its image-encoder
weights initialize the Query Image Encoder. The pretraining latent
and confidence heads are discarded. The Ordered Preference Encoder, Query Image
Encoder, and Identity-Residual LUT Decoder remain frozen during personalized training.

\noindent\textbf{Personalized Training.}
The Preference Set Aggregator, LUT latent vector head, and edit-strength head are
randomly initialized and optimized during personalized training. The standard
identity profile mappings have no trainable parameters. The objective weights
$(\lambda_c,\lambda_r,\lambda_a,\lambda_p,\lambda_g,\lambda_l,\lambda_u)$ are
$(1,1,2,3,0.5,0.05,0.5)$, with ranking margin $m=0.02$ and $\tau=0.05$ in the
color-distance loss denominator. The model is trained for 100 epochs using
128-pixel crops, batch size 16, four query pairs per user, and AdamW with learning
rate $2\times10^{-4}$, cosine decay, weight decay $10^{-4}$, and gradient clipping at 1.0.

The five checkpoints with the lowest fixed validation loss are evaluated on the
full validation episode. Epoch 91 gives the highest CQS on all four measures
among these candidates. PrefLUT uses this epoch-91 checkpoint for PPSD quality,
PCVP, and profile refinement. PPSD quality results are validation
estimates after checkpoint selection on the same users and evaluation episode.

\noindent\textbf{Inference.}
The main PPSD quality results use training seed 2026 and an evaluation episode
constructed with sampling seed 2028.
Query construction and image processing follow Appendix~\ref{app:pcvp-settings}.
The effective inference strength in Equation~\ref{eq:gatedlut} is
$g^{\mathrm{eval}}_{u,q}=0.7g_{u,q}$. The residual scale of 0.7 is fixed across all
users, control conditions, and five candidate checkpoints. Table~\ref{tab:adapted-personalized}(A,B) uses unquantized FP32
profiles. The int8 deployment and timing configuration is specified in
Appendix~\ref{app:efficiency-settings}.

\subsection{Baseline Reproduction and Adaptation}
\label{app:baseline-reproduction}

The following descriptions follow the baseline order in Table~\ref{tab:adapted-personalized}.
All eight baseline adaptations use ten PPSD training epochs and report the
final-epoch checkpoint without validation-based checkpoint selection. PieNet also
receives ten epochs of triplet pretraining before this adaptation stage.
PieNet, StarEnhancer, PIE-MSM, DiffRetouch, and PerTouch are adapted to PPSD
using training seed 2026. Training uses
471 users, 16 reference pairs, and four query pairs per user. Reference and query
pairs are disjoint. The shared pairwise objective assigns weights $(1,1,2,3)$ to
the color-distance, ranking, aligned-pair reconstruction, and preservation losses.
Evaluation uses 50 validation users, $N=M=16$, and episode seed 2028, giving
800 query pairs. For the Cyclic Wrong-Query Control, the conditioning features
below are computed from another query image of the same user.

\noindent\textbf{PieNet.}
The PieNet adaptation~\citep{kim2020pienet} uses a ResNet-18 encoder and randomly
initialized personalized layers. The model first receives triplet pretraining using reference images, followed by pairwise training
with Adam at learning rate $10^{-4}$ and 256-pixel images. For each new user,
a normalized 512-dimensional user code is fitted to 16 preference pairs while
the networks remain frozen. Fitting uses a triplet margin of 0.2 and 100 Adam
steps at learning rate 0.05. The Cyclic Wrong-Query Control replaces the condition
at the global bottleneck while keeping skip-connection features from the original
query.

\noindent\textbf{StarEnhancer.}
StarEnhancer~\citep{song2021starenhancer} starts from public pretrained weights.
The mapping network and enhancer are updated while the style encoder remains
frozen. Training uses Adam at learning rate $10^{-5}$, 256-pixel images, and
gradient clipping at 1.0. Non-preferred and preferred reference
features are aggregated separately to form the source and target style centers.
The Mismatched-Pair Profile preserves both centers and is therefore marked as
not applicable. The Cyclic Wrong-Query Control replaces the query features used
to predict enhancement parameters.

\noindent\textbf{PIE-MSM.}
PIE-MSM~\citep{kosugi2023masked} starts from public pretrained weights. Training
updates the Transformer, masked-style and content modules, and enhancer using
Adam at learning rate $10^{-5}$, 512-pixel images, and gradient clipping at 1.0. Each reference token combines the preferred-minus-non-preferred
style feature difference with the content feature of the non-preferred image.
The Cyclic Wrong-Query Control replaces the query image's content features supplied to the
Transformer while keeping the original image as the editing input.

\noindent\textbf{DiffRetouch.}
DiffRetouch~\citep{duan2024diffretouch} starts from public weights and updates
the U-Net and HDR decoder while the VAE remains frozen. Training uses AdamW at
learning rate $10^{-6}$ and weight decay 0.01. A denoising loss
with weight one is added to the shared pairwise objective for preferred-input
preservation and aligned-pair reconstruction. Quality and PCVP evaluation use 20 sampling steps. The method's attribute measurements
produce a four-dimensional control vector from 16 reference pairs. For each
attribute, the sign of the preferred-minus-non-preferred difference is averaged
over the pairs. The attributes are colorfulness, brightness, contrast, and
temperature, in that order. The Cyclic Wrong-Query Control replaces the
low-resolution image condition. Bilateral-grid guidance and application use the original query.

\noindent\textbf{PerTouch.}
The PerTouch adaptation~\citep{chang2026pertouch} starts from public weights and
updates the ControlNet and output convolution while the VAE remains frozen.
Training uses AdamW at learning rate $10^{-6}$, weight decay 0.01,
and the shared pairwise objective. A denoising loss with weight one is added for
preferred-input preservation and aligned-pair reconstruction. Quality and PCVP
evaluation use 50 sampling steps. Attribute measurements
produce a four-dimensional control vector by averaging the signs of
preferred-minus-non-preferred differences over 16 pairs. The attribute order is
colorfulness, contrast, temperature, and brightness. The vector is expanded to
a spatially constant parameter map for the public diffusion backbone. This
adaptation excludes the VLM Agent's semantic masks and editing memory.
The Cyclic Wrong-Query Control replaces the VAE image condition. The architecture
has no separate path that applies a predicted transform to the original query.
This control measures dependence on the VAE image condition, including image
content, rather than isolating color-transform prediction on a fixed image.
For both diffusion methods, the Training-User Mean Profile averages profiles from
471 training users. Correct and controlled conditions share diffusion seed
$2028+32i+2j+b$, where $i$ is the user index, $j$ is the query-pair index, and
$b\in\{0,1\}$ identifies the preferred or non-preferred input. All indices start at zero.

\noindent\textbf{User-specific Decoder.}
The implementation follows the EDSR/HIIF mechanism in \citet{kim2026ppsd} and is
trained on PPSD. For a new user, the shared image encoder remains frozen
while decoder weights are fitted to the user's reference pairs using 100 Adam
steps at learning rate $10^{-4}$. This implementation uses the released
image-processing components and supplies the quality, PCVP, and efficiency
results. The Training-User Mean Profile is formed by averaging decoder weights
fitted to the reference sets of 471 training users.

\noindent\textbf{User Preference Embedding (UPE).}
The UPE adaptation~\citep{kim2026ppsd} combines a frozen DINOv2 ViT-S/14 encoder
with a locally implemented enhancement network. Training uses the shared pairwise
loss and Adam at learning rate $10^{-4}$. This model supplies the quality results
in Table~\ref{tab:adapted-personalized}(A,B) and the PCVP results. The Cyclic Wrong-Query Control replaces spatial encoder
features while keeping the original query's RGB values, coordinates, and residual
skip connections. The efficiency results in Table~\ref{tab:adapted-personalized}(C)
use a separate, untrained EDSR/HIIF model with a 256-dimensional preference embedding and a pretrained frozen
DINOv2 encoder. For this timing model, the preference embedding is concatenated
with the input to the first HIIF MLP. Preference fusion uses two layers and eight attention heads.

\noindent\textbf{Exemplar-based Inference.}
The implementation follows \citet{kim2026ppsd} and adapts PIE-MSM
using triplets of scene, preferred, and non-preferred images. A learned projection
maps the resulting 1,536-dimensional triplet features to 1,024 dimensions.
This implementation uses the released image-processing components and supplies
the quality, PCVP, and efficiency results. The Reversed-Order Profile and
Mismatched-Pair Profile retain the original scene and content features.

\subsubsection{Personalized Enhancement Methods Not Compared on PPSD}
\label{app:uncompared-personalized}

\noindent\textbf{RefRetouch.}\quad
RefRetouch~\citep{weldengus2026refretouch} combines a LoRA-adapted Siamese
SigLIPv2 encoder, a conditional color-space MLP, and retrieval of reference
retouching latents. It learns from original/retouched image pairs. Its project
page lists ECCV 2026 and marks the code as ``to be released'' as of
19 September 2026.\footnote{\url{https://tememuruts.github.io/RefRetouch-Project-Page/}}
PPSD comparison requires adapting the reference interface and training objective
to ordered preferences that need not be pixel-aligned. RefRetouch is therefore
discussed but excluded from quantitative comparisons.

\noindent\textbf{FedPAIE.}\quad
The 2026 preprint FedPAIE~\citep{xu2026fedpaie} studies federated learning from
private scalar aesthetic ratings. A locally calibrated scorer guides per-user
adaptation of a CLUT enhancer. Calibration combines rating regression, pairwise
ranking derived from ratings, and a rating-variance objective. PPSD supplies
relative choices without the scalar targets required by these objectives.
Adapting FedPAIE would therefore require redefining its scorer supervision and
client training protocol, in addition to the enhancement objective. Its
decentralized data constraints and scorer-guided adaptation thus place it outside
the current PPSD comparison.

\subsection{Implementation Details for Extended Experiments}
\label{app:comparison-settings}
\label{app:task-settings}

FiveK uses Expert C, and PPR10K uses Experts A, B, and C to evaluate expert-target
enhancement. Dataset splits and processing are specified in
Appendices~\ref{app:fivek-dataset} and~\ref{app:ppr10k-dataset}.

\noindent\textbf{Color Curves, LUT, and Spatial Residual.}\quad
Both configurations concatenate the expanded user feature and query feature as
$\xi=[\bar p_u,c_q]\in\mathbb R^{512}$. Two heads, each using LayerNorm,
a $512\rightarrow512$ linear layer and SiLU, predict 48 curve logits and 64 LUT
coefficients, respectively. For channel $c$, the 16 logits $v_{c,k}$ define
positive increments and 17 monotone knots:
\[
 a_{c,k}=\operatorname{softplus}(v_{c,k})+10^{-6},\qquad
 q_{c,0}=0,\qquad
 q_{c,j}=\frac{\sum_{k=1}^{j}a_{c,k}}{\sum_{k=1}^{16}a_{c,k}}.
\]
Linear interpolation between knots at input values $j/16$ defines the channelwise
curve $C_\xi$. Let $B_0$ and $\{B_k\}_{k=1}^{64}$ be learned residual grids of
size $17^3\times3$, and let $b_k(\xi)$ be the coefficient-head outputs. The
query-conditioned LUT is:
\[
 L_\xi=\operatorname{clip}_{[0,1]}\!\left(L_{\mathrm{id}}+
 0.5\tanh\!\left(B_0+\sum_{k=1}^{64}\tanh(b_k(\xi))B_k\right)\right).
\]
The rank refers to the 64 learned residual bases. For query image $I$, the global output is
$J=\mathcal T(L_\xi,C_\xi(I))$. A spatial network takes the six-channel
concatenation $[I,J]$ resized to half resolution. Its three $3\times3$ convolutions
have channels $6\rightarrow24\rightarrow24\rightarrow3$, with SiLU after the
first two. After the first SiLU, the feature map $F$ is modulated as
$(1+\gamma)\odot F+\delta$, where a linear map of $\xi$ predicts the channelwise
scale $\gamma$ and shift $\delta$. Bilinear upsampling restores
the spatial residual $R_\xi(I,J)$ to the original resolution. Both configurations
use full edit strength, giving the final image:
\[
 \widehat I=\operatorname{clip}_{[0,1]}\bigl(J+0.25\tanh(R_\xi(I,J))\bigr).
\]
Spatial resizing uses bilinear interpolation without aligned corners. These
components are trained jointly for the expert targets. The additional spatial
residual is applied after the global color transform and remains a separate
image-processing component.

\subsubsection{FiveK Expert-Target Enhancement}
\label{app:fivek-settings}
The 6.127M-parameter FiveK configuration uses the components above. Its Reference
Image Encoder and compact Query Image Encoder are trained from scratch. Each of
the eight training runs uses batch size 32,
Adam with learning rate $3\times10^{-4}$, no weight decay, warmup
followed by cosine decay, and gradient clipping at 1.0.
References are resized to 128 pixels. Training uses 384-pixel crops, horizontal
flips, and brightness and saturation augmentation. The architecture and training
duration are selected using the validation subset of the training images, and each final checkpoint
is evaluated on all 500 test images. All eight training runs contribute equally
to the mean.

\subsubsection{PPR10K Expert-Target Enhancement}
The 17.156M-parameter PrefLUT configuration jointly trains a shared model with
monotone color curves, a rank-64 $17^3$ LUT, an ImageNet-pretrained ResNet-18 Query
Image Encoder, and a spatial residual. The Reference Image Encoder and Preference
Set Aggregator are randomly initialized, and no PPSD enhancement checkpoint is
reused. Each expert profile is constructed from the same 16 training images, with
the expert-retouched target as preferred and the source as non-preferred.

Architecture selection uses 8,005 training images and an internal validation subset
of 870 images, with no shared image groups, and compares the epoch-8 checkpoints.
The selected architecture is then trained on all 8,875 training images for eight
epochs using seeds 2034 and 2036. References are resized to 128 pixels and training
inputs to 224 pixels. Training uses batch size 12, AdamW at $3\times10^{-4}$,
cosine decay, weight decay $10^{-4}$, and gradient clipping at 1.0. The final epoch-8
checkpoint of each run is evaluated against each expert's targets for the same
2,286 source images in the released 360p validation set~\citep{liang2021ppr10k}.
Results average both training runs equally.

\subsection{Efficiency Measurement}
\label{app:efficiency-settings}

PrefLUT latency measurements and the profile-reuse benchmark use FP32.
Baseline latency measurements use BF16 automatic mixed precision.
Latency measurements disable TF32.

\noindent\textbf{Main Comparison.}
All methods use one RTX~5090 and four CPU threads on a Ryzen~9~9950X3D.
All methods use the same eight PPSD validation users, 32 query images at $512\times512$, and
16 reference pairs per user from episode seed 2028. Users are sorted by identifier,
and the first eight are selected. Both images from each user's first two sampled
query pairs are timed. Query batch size is one.
After one profile-construction warmup and five editing warmups, latency is the
mean elapsed time measured with GPU synchronization. Loading, external resizing,
input transfer, serialization, and quality metrics are excluded. All internal
image transforms, LUT application, and decoding are timed.

Reference features are extracted one image at a time at $224\times224$ for DINOv2
and $256\times256$ otherwise. Construction includes both images of each pair,
PieNet's user-code fitting, PIE-MSM's style and content encoding, and UPE's DINOv2
features. UPE uses the untrained HIIF implementation specified in
Appendix~\ref{app:baseline-reproduction}. User-specific Decoder construction includes 100 Adam fitting steps
at $10^{-4}$. Exemplar-based Inference encodes the scene, preferred, and non-preferred reference images. DiffRetouch and
PerTouch compute their image attributes on the CPU and use 20 and 50 sampling steps,
respectively.

Parameters include the complete networks needed by either stage,
including frozen backbones. Profile storage follows each method's saved representation.
User-specific Decoder stores its fitted decoder parameters per user.
For PrefLUT, Table~\ref{tab:adapted-personalized}(A,B) reports the epoch-91 model
with an unquantized 256-dimensional FP32 profile, occupying 1,024 bytes per user.
Panel (C) measures a separate checkpoint with the same architecture and an int8
profile: 256 bytes for its entries plus one four-byte FP32 scale, totaling
260 bytes per user. Profile quantization and dequantization are included in the
timed stages.

The computational cost includes convolutions and matrix multiplications needed
to edit one query using a stored profile. Each multiply-add counts as two FLOPs. Attention and all diffusion
steps are included. Grid sampling, resizing, normalization, softmax, activations,
and other elementwise operations are excluded, making these counts lower bounds
on total arithmetic. The same counting rule applies to every method.

\noindent\textbf{Profile Reuse Benchmark.}
A separate benchmark uses two users with 16 reference pairs and 16 query pairs
per user, giving 64 query images per batch. The benchmark covers reference encoding
and parameter prediction, excluding LUT decoding, image application, and profile
quantization. References and queries are resized
to 256 and 512 pixels, respectively. Timing uses three warmup iterations followed
by 500 measured repetitions. Predicting LUT latent vectors and edit strengths
takes 77.204236\,ms when reference sets are encoded separately for all 64 queries
and 8.437795\,ms when the two Reusable User Profiles are reused. The ratio is
$77.204236/8.437795=9.149812\approx9.150$.
Initial reference encoding costs 2.131532\,ms, giving 10.569326\,ms for the first
batch and a $7.305\times$ speedup when that cost is included. Table~\ref{tab:adapted-personalized}(C)
instead measures complete single-query editing with a stored profile.

\section{Preference-Conditioning Verification Protocol (PCVP)}
\label{app:pcvp}

PCVP assesses whether collecting and storing a user's preferences improves editing
for that user. An editor can produce attractive images using an enhancement rule
shared across users, so image quality alone does not establish the value of personal
feedback. PCVP compares edits of the same query under controlled changes to
preference and query conditions. The resulting gains provide a common basis for
verifying that the supplied preferences and current image contribute to editing
quality. The controls, pass criteria, and PPSD settings are detailed below.

\subsection{Detailed PCVP Controls}
\label{app:pcvp-details}

PCVP tests whether editing quality benefits from the intended user's profile,
preference order, pair correspondence, individual preferences, and the query image.
Each test changes one condition while keeping the shared model parameters,
evaluated users, query-target pairs,
image-processing steps, and metric computation fixed. The four profile
controls keep the query condition unchanged. The Cyclic Wrong-Query Control
keeps the user profile unchanged.

Let $S_u=\{(I_i^+,I_i^-)\}_{i=1}^{N_u}$ be user $u$'s ordered preference image pairs,
where $I_i^+$ and $I_i^-$ are the preferred and non-preferred images, and let
$G$ denote the evaluated method's profile construction function, so $p_u=G(S_u)$. The function
$F(p,Q_{\mathrm{cond}},Q_{\mathrm{apply}})$ denotes the output under profile $p$
and conditioning query $Q_{\mathrm{cond}}$, retaining $Q_{\mathrm{apply}}$ on any
separate image-application path.
Under the correct condition, $Q_{\mathrm{cond}}=Q_{\mathrm{apply}}$ is the current query image. Method-specific
conditioning interfaces are given in Appendix~\ref{app:baseline-reproduction}.

\subsubsection{Fixed Wrong-User Profile}

Users can prefer different brightness, contrast, or color for the same photo.
This test asks whether using the intended user's profile improves the result
compared with another user's profile. It evaluates whether stored profiles preserve
user differences that matter for editing.

A fixed user reassignment $\rho_U$ replaces $p_u$ with another user's profile
$p_{\rho_U(u)}$, with $\rho_U(u)\neq u$. On PPSD, each user receives the next user's
profile in a fixed order, and the last user receives the first user's profile.
Each replacement profile is built from the other user's correct reference set.
The evaluated user's query images and targets remain fixed. A positive CQS gain
for the correct profile therefore shows the benefit of the intended user's preferences.

\subsubsection{Reversed-Order Profile}

Pairwise feedback records the preferred alternative. Choosing the darker or
brighter version of the same pair should provide different editing guidance.
This test checks whether the editor follows the user's choice and whether
preference labels affect editing quality.

The preferred and non-preferred images are exchanged within every reference pair:
\begin{equation}
  S_u^{\mathrm{rev}}=\{(I_i^-,I_i^+)\}_{i=1}^{N_u},\qquad
  p_u^{\mathrm{rev}}=G(S_u^{\mathrm{rev}}).
  \label{eq:pcvp-reversed-app}
\end{equation}
The reference images, pair membership, and reference count remain unchanged.
Only preference direction changes. A positive gain supports the use of ordered
choices beyond an unordered collection of reference appearances.

\subsubsection{Mismatched-Pair Profile}

Each choice is made against a specific alternative. Choosing a warm image over
a cooler image conveys a different comparison from choosing it over a more
saturated image. This test assesses whether preserving each original pair
improves the resulting edits.

For methods using paired references, $\rho_P(i)=1+(i\bmod N_u)$ cyclically shifts
the reference-pair indices. Each preferred image is paired with the next
non-preferred image:
\begin{equation}
  S_u^{\mathrm{mis}}=\{(I_i^+,I_{\rho_P(i)}^-)\}_{i=1}^{N_u},\qquad
  p_u^{\mathrm{mis}}=G(S_u^{\mathrm{mis}}).
  \label{eq:pcvp-mismatched-app}
\end{equation}
The reassignment preserves the complete preferred and non-preferred image sets
and the number of reference pairs, but breaks the original pair correspondence
within each user. A positive CQS gain shows the benefit of that correspondence
beyond separately aggregating the two image sets. The Reversed-Order Profile
changes which image is preferred within each pair. The Mismatched-Pair Profile
changes which images are paired.

\subsubsection{Training-User Mean Profile}

An editing service could apply one default profile to every user. Here, that
default is the mean profile of training users. This test asks whether an
individual's feedback provides better edits than the shared default, evaluating
the value of collecting personal feedback and maintaining individual profiles.

One profile is constructed from a fixed reference set for each training user, then
averaged with equal user weights:
\begin{equation}
  \overline p_{\mathrm{train}}=
  \frac{1}{|\mathcal U_{\mathrm{train}}|}
  \sum_{v\in\mathcal U_{\mathrm{train}}}G(S_v),\qquad
  p_u^{\mathrm{mean}}=\overline p_{\mathrm{train}}.
  \label{eq:pcvp-mean-app}
\end{equation}
Only training users contribute to the mean profile. When a profile contains a
sequence of features, the features are first averaged within each training user,
then across users. The averaging rules for User-specific Decoder, DiffRetouch, and
PerTouch are specified in Appendix~\ref{app:baseline-reproduction}.

Every evaluated user's profile is replaced by the same training-user mean. A positive CQS gain shows the benefit
of individual preferences over the shared population profile. The Fixed Wrong-User
Profile instead compares the intended user's preferences with another individual's preferences.

\subsubsection{Cyclic Wrong-Query Control}

Applying the same preference to photos with different exposures and color
distributions can require different transforms. For example, a transform predicted
for a dark indoor image may be unsuitable for a bright outdoor image. This test
checks whether reusable preferences are translated into edits suited to each
current photo.

For PrefLUT, query pairs are sorted by pair identifier and cyclically shifted
within each user. Preferred and non-preferred inputs use the corresponding image
from the next pair as their query condition. Here $\rho_Q(q)$ indexes the corresponding
conditioning image for query $q$, with $\rho_Q(q)\neq q$. The original query is retained
on any separate image-application path:
\begin{equation}
  \widehat I_{u,q}^{\mathrm{wrongQ}}
  =F(p_u,Q_{u,\rho_Q(q)},Q_{u,q}),\qquad
  \widehat I_{u,q}^{(0)}=F(p_u,Q_{u,q},Q_{u,q}).
  \label{eq:pcvp-query-app}
\end{equation}
The user profile and evaluation target remain fixed. In PrefLUT, the substituted
query determines the predicted LUT and edit strength. Both transforms are applied
to the same original image, so a positive CQS gain shows the benefit of using the
current query to predict its color transform. PerTouch instead replaces its VAE
image condition. Its gain also reflects image-content dependence
(Appendix~\ref{app:baseline-reproduction}).

\subsection{Evaluation Criterion and PPSD Settings}
\label{app:pcvp-settings}

\noindent\textbf{Pass Criterion.}
The gain is correct-condition CQS minus control-condition CQS
(Equation~\ref{eq:pcvp-gain}). On PPSD, CQS is computed separately for
$\Delta E_{00}$, LPIPS, PSNR, and SSIM, with higher scores indicating better
quality. Positive gains favor the correct condition. A change in output alone
does not show better quality.

The paired bootstrap resamples users while retaining all queries of each selected
user and the pairing between conditions. A metric passes when the lower endpoint of
its paired 95\% interval is strictly positive. For repeated evaluation episodes,
the gain must also retain its direction as required by the main protocol.
Bootstrap procedures are given in Appendix~\ref{app:metric-statistics}, with
control-specific settings below.

For each control, the pass count is the number of metrics with a statistically
positive CQS gain, ranging from 0/4 to 4/4. This count measures the evidence for
effective conditioning rather than absolute image quality. A full PCVP pass requires
4/4 for all five controls, as defined in Equation~\ref{eq:pcvp-criterion}.
PCVP changes the profile or query condition while keeping shared model parameters fixed. Architecture ablations instead
compare models trained with different components.

\noindent\textbf{PPSD Evaluation Settings.}
The eight baseline adaptations are evaluated with 50 validation users, 16 reference
pairs and 16 query pairs per user, and episode seed 2028. Reference images are
resized to 256 pixels and query images to 512 pixels. Metrics are computed at
256 pixels. Each baseline control uses the same model checkpoint and sampled
reference and query sets as its correct condition. Confidence intervals use
250,000 paired user-bootstrap resamples.

PrefLUT's Fixed Wrong-User Profile test uses the same episode and 512-pixel
queries as the quality evaluation. Its confidence intervals use 250,000 paired
user-bootstrap resamples with bootstrap seed 2028.
The Reversed-Order Profile and Mismatched-Pair Profile tests use 256-pixel inputs
and 100,000 resamples. The Training-User Mean Profile and Cyclic Wrong-Query Control
tests apply transforms at the original image resolution and compute metrics at
256 pixels, using 250,000 resamples. PPSD training uses seeds 2026 and 2027.
The main comparison reports training seed 2026 and evaluation episode seed 2028.

\noindent\textbf{Repeated Evaluation Episodes.}
For the fixed model, three additional episodes use seeds 2029 to 2031 to repeat
the Fixed Wrong-User Profile, Reversed-Order Profile, and Cyclic Wrong-Query Control
tests. These episodes keep the validation users, reference and query counts,
image-processing steps, metrics, and control rules unchanged.

\section{Additional Experimental Results}
\label{app:baselines}

\subsection{PPSD PCVP Results}
\label{app:baseline-pcvp-complete}

\noindent\textbf{Control Applicability.}
StarEnhancer constructs source and target style centers by separately averaging
non-preferred and preferred reference features, then normalizing the two averages.
The Mismatched-Pair Profile changes pair correspondence but preserves both image
sets, so neither style center changes. This control therefore leaves StarEnhancer's
condition unchanged and is marked as not applicable in Table~\ref{tab:pcvp-main}.
The conditioning interface determines whether a control applies, independently of
the measured gain. StarEnhancer passes one of its four applicable controls.
The reported 1/4 counts applicable controls, whereas full PCVP requires passing all five.

\noindent\textbf{Baseline Conditioning Patterns.}
Table~\ref{tab:pcvp-main} distinguishes dependence on the query image from the
benefit of an individual user's preferences. StarEnhancer, DiffRetouch, and PerTouch
pass all four metrics in the Cyclic Wrong-Query Control, but none passes all four
in either the Fixed Wrong-User Profile or Training-User Mean Profile test.
Thus, their use of the current image does not establish a consistent quality
advantage for the intended user's profile. In the two diffusion adaptations,
references are reduced to four averaged attribute-sign values, which can map
different reference sets to similar controls. PerTouch's query intervention also
changes its VAE image condition, so its large query gains include image-content
dependence, as explained in Appendix~\ref{app:baseline-reproduction}.

PIE-MSM passes the Reversed-Order Profile, Mismatched-Pair Profile, and Cyclic
Wrong-Query Control tests, consistent with its joint use of style-difference and content features. However,
its correct profile does not improve all four metrics over another user's profile
or the training-user mean. These results support its use of structured reference
evidence while leaving the stronger user-specific benefit unverified by the full
PCVP criterion. PieNet, UPE, User-specific Decoder, and Exemplar-based Inference
show gains on selected metrics or controls without passing any complete test.
For PieNet and UPE, the query intervention retains original-image paths, so weak
query gains concern the replaced conditioning features rather than dependence on
the input image as a whole. These conclusions apply to the evaluated PPSD
adaptations and their conditioning interfaces, without ruling out personalization
in the original methods.

\subsubsection{PrefLUT Stability across Evaluation Episodes}
\label{app:multiepisode}

We test the stability of PrefLUT's PCVP results across repeated evaluation sampling,
keeping the trained checkpoint and validation users fixed. We repeat the Fixed
Wrong-User Profile, Reversed-Order Profile, and Cyclic Wrong-Query Control tests
with episode seeds 2029 to 2031, in addition to the seed-2028 evaluation in
Table~\ref{tab:pcvp-main}. This experiment measures sensitivity to evaluation
sampling, rather than variation across independently trained models.
For PrefLUT, combining the three episodes yields positive gains supported by the
PCVP criterion for all four metrics in each test. The Fixed Wrong-User Profile
test passes three, three, and four metrics in the individual episodes,
respectively. The other two tests pass all four metrics in every episode.
Table~\ref{tab:multiepisode-confirmation} reports these combined PrefLUT results.

\begin{table}[!htbp]
  \centering
  \caption{PrefLUT PCVP gains across evaluation episodes on PPSD. Gains are
  correct-condition CQS minus the named control's CQS, combined over sampling seeds
  2029--2031 with the model and users fixed. Brackets give paired 95\%
  user-bootstrap confidence intervals.}
  \label{tab:multiepisode-confirmation}
  \fontsize{8.5}{10}\selectfont
  \setlength{\tabcolsep}{1pt}
  \renewcommand{\arraystretch}{1.2}
  \setlength{\AppMetricWidth}{\dimexpr(\textwidth-.27\textwidth-10\tabcolsep-\arrayrulewidth)/4\relax}
  \begin{tabular}{>{\raggedright\arraybackslash}m{.27\textwidth}|>{\columncolor{AppBlueA}\centering\arraybackslash}m{\dimexpr\AppMetricWidth-3pt\relax}>{\columncolor{AppBlueB}\centering\arraybackslash}m{\dimexpr\AppMetricWidth+9pt\relax}>{\columncolor{AppBlueA}\centering\arraybackslash}m{\dimexpr\AppMetricWidth-3pt\relax}>{\columncolor{AppBlueB}\centering\arraybackslash}m{\dimexpr\AppMetricWidth-3pt\relax}}
    \toprule
  \mbox{PCVP Test} & \cellcolor{AppBlueHeader}$\Delta E_{00}\uparrow$ & \cellcolor{AppBlueHeader}LPIPS $\uparrow$ & \cellcolor{AppBlueHeader}PSNR $\uparrow$ & \cellcolor{AppBlueHeader}SSIM $\uparrow$ \\
    \midrule
    
  \mbox{Fixed Wrong-User Profile} & \appgain{\shortstack{$0.003999$\\$[\!.002139,\!.006048]$}} & \appgain{\shortstack{$0.412528$\\$[\!.194432,\!.698803]$}} & \appgain{\shortstack{$0.267460$\\$[\!.036182,\!.512614]$}} & \appgain{\shortstack{$0.001102$\\$[\!.000600,\!.001662]$}} \\[5pt]
  \mbox{Reversed-Order Profile} & \appgain{\shortstack{$0.006449$\\$[\!.003358,\!.010141]$}} & \appgain{\shortstack{$0.383434$\\$[\!.109178,\!.764762]$}} & \appgain{\shortstack{$0.612272$\\$[\!.265722,\!.991219]$}} & \appgain{\shortstack{$0.001116$\\$[\!.000539,\!.001746]$}} \\[5pt]
  \mbox{Cyclic Wrong-Query Control} & \appgain{\shortstack{$0.015708$\\$[\!.012523,\!.019271]$}} & \appgain{\shortstack{$2.393584$\\$[\!1.629063,\!3.454507]$}} & \appgain{\shortstack{$0.881433$\\$[\!.773038,\!.990171]$}} & \appgain{\shortstack{$0.005383$\\$[\!.004403,\!.006354]$}} \\[5pt]
    \bottomrule
  \end{tabular}
\end{table}

\subsection{Enhancement on FiveK and PPR10K}
\label{app:transfer-comparisons}

Tables~\ref{tab:fivek-personalized} and~\ref{tab:ppr10k-preflut} include published
results for SpliNet~\citep{bianco2020splinet}, PieNet, StarEnhancer, and PIE-MSM.
Their values come from Table~I of \citet{kosugi2023masked}, using 100 reference
pairs per new expert and averaging ten reference-set samples. The experts are
unseen during model training. PrefLUT uses 16 reference pairs and is trained
toward the evaluated expert targets (Appendix~\ref{app:task-settings}). These
settings differ in training data, expert coverage, and image resolution, so the
tables separate the protocols without a shared ranking.

\subsubsection{MIT-Adobe FiveK}
\label{app:fivek-results}
Table~\ref{tab:fivek-personalized}(A) averages the published personalization
results across Experts A/B/C/D/E. Panel (B) evaluates PrefLUT
on the 500-image Expert-C test set across eight training seeds. Its 480-pixel
evaluation preserves aspect ratio and uses Wang-2009 automatic downsampling
for SSIM (Appendix~\ref{app:fivek-dataset}). The published protocol uses
$512\times512$ inputs to the enhancer.

\begin{table}[!htbp]
  \centering\small
  \setlength{\tabcolsep}{4pt}
  \renewcommand{\arraystretch}{1.08}
  \caption{MIT-Adobe FiveK results under different evaluation protocols.
  (A) Published unseen-expert personalization results averaged over Experts A/B/C/D/E.
  (B) PrefLUT trained for Expert C and evaluated on held-out images, reported as
  mean $\pm$ sample standard deviation over 8 training seeds.}
  \label{tab:fivek-personalized}
  \setlength{\AppMetricWidth}{\dimexpr(\textwidth-.34\textwidth-8\tabcolsep-1\arrayrulewidth)/3\relax}
  \begin{tabular}{L{.34\textwidth}|>{\columncolor{AppOrangeA}}w{c}{\AppMetricWidth}>{\columncolor{AppOrangeB}}w{c}{\AppMetricWidth}>{\columncolor{AppOrangeA}}w{c}{\AppMetricWidth}}
  \toprule
  \multicolumn{4}{c}{\cellcolor{AppOrangeHeader}(A) Unseen-Expert Personalization, 100 Reference Pairs} \\
  \midrule
  Method & PSNR $\uparrow$ & SSIM $\uparrow$ & $\Delta E_{ab}$ $\downarrow$ \\
  \midrule
  SpliNet (TIP 2020) & 19.94 & 0.840 & 13.87 \\
  PieNet (ECCV 2020) & 20.54 & 0.851 & 13.56 \\
  StarEnhancer (ICCV 2021) & 19.68 & 0.833 & 14.94 \\
  PIE-MSM (TCSVT 2024) & 23.18 & 0.898 & 10.13 \\
  \bottomrule
  \end{tabular}\par\nointerlineskip
  \setlength{\AppMetricWidth}{\dimexpr(\textwidth-.22\textwidth-10\tabcolsep-1\arrayrulewidth)/4\relax}
  \begin{tabular}{L{.22\textwidth}|*{2}{>{\columncolor{AppBlueA}}w{c}{\AppMetricWidth}>{\columncolor{AppBlueB}}w{c}{\AppMetricWidth}}}
  \multicolumn{5}{c}{\cellcolor{AppBlueHeader}(B) Expert-C Enhancement, 16 Reference Pairs} \\
  \midrule
  Method & PSNR $\uparrow$ & SSIM $\uparrow$ & $\Delta E_{76}$ $\downarrow$ & Alex-LPIPS $\downarrow$ \\
  \midrule
  PrefLUT & $24.5717\pm0.0815$ & $0.91718\pm0.00060$ & $7.8331\pm0.0629$ & $0.06663\pm0.00054$ \\
  \bottomrule
  \end{tabular}
\end{table}

\subsubsection{PPR10K}
\label{app:ppr10k-results}

Table~\ref{tab:ppr10k-preflut}(A) averages published personalization results
across Experts A/B/C. Panel (B) evaluates a shared PrefLUT model
trained on all three expert targets. Both settings use 2,286 evaluation images
per expert. PrefLUT uses the 360-pixel release, while the published protocol
uses $512\times512$ enhancer inputs. The PrefLUT results average two training
seeds and include human-centered metrics that give greater weight to regions
covered by the released human masks (Appendix~\ref{app:ppr10k-dataset}).

\begin{table}[!htbp]
  \centering\small
  \setlength{\tabcolsep}{4pt}
  \renewcommand{\arraystretch}{1.08}
  \caption{PPR10K results under different evaluation protocols.
  (A) Published unseen-expert personalization results averaged over Experts A/B/C.
  (B) One PrefLUT model trained jointly for these experts, with each row averaged
  over 2 training seeds. HC denotes human-centered metrics.}
  \label{tab:ppr10k-preflut}
  \setlength{\AppMetricWidth}{\dimexpr(\textwidth-.34\textwidth-8\tabcolsep-1\arrayrulewidth)/3\relax}
  \begin{tabular}{L{.34\textwidth}|>{\columncolor{AppOrangeA}}w{c}{\AppMetricWidth}>{\columncolor{AppOrangeB}}w{c}{\AppMetricWidth}>{\columncolor{AppOrangeA}}w{c}{\AppMetricWidth}}
  \toprule
  \multicolumn{4}{c}{\cellcolor{AppOrangeHeader}(A) Unseen-Expert Personalization, 100 Reference Pairs} \\
  \midrule
  Method & PSNR $\uparrow$ & SSIM $\uparrow$ & $\Delta E_{ab}$ $\downarrow$ \\
  \midrule
  SpliNet (TIP 2020) & 20.34 & 0.865 & 12.84 \\
  PieNet (ECCV 2020) & 19.97 & 0.824 & 14.32 \\
  StarEnhancer (ICCV 2021) & 21.22 & 0.885 & 12.29 \\
  PIE-MSM (TCSVT 2024) & 22.97 & 0.921 & 9.83 \\
  \bottomrule
  \end{tabular}\par\nointerlineskip
  \setlength{\AppMetricWidth}{\dimexpr(\textwidth-.22\textwidth-10\tabcolsep-1\arrayrulewidth)/4\relax}
  \begin{tabular}{L{.22\textwidth}|*{2}{>{\columncolor{AppBlueA}}w{c}{\AppMetricWidth}>{\columncolor{AppBlueB}}w{c}{\AppMetricWidth}}}
  \multicolumn{5}{c}{\cellcolor{AppBlueHeader}(B) Enhancement for Experts A/B/C, 16 Reference Pairs} \\
  \midrule
  Target Expert & \cellcolor{AppBlueHeader}PSNR $\uparrow$ & \cellcolor{AppBlueHeader}$\Delta E_{76}$ $\downarrow$ & \cellcolor{AppBlueHeader}PSNR-HC $\uparrow$ & \cellcolor{AppBlueHeader}$\Delta E_{76}$-HC $\downarrow$ \\
  \midrule
  Expert A & 25.2211 & 7.5897 & 28.4737 & 4.9233 \\
  Expert B & 24.8876 & 7.9971 & 28.1469 & 5.1686 \\
  Expert C & 25.2069 & 7.8780 & 28.4842 & 5.1150 \\
  \midrule
  \cellcolor{AppBlueHeader}Mean over A/B/C & 25.1052 & 7.8216 & 28.3683 & 5.0690 \\
  \bottomrule
  \end{tabular}
\end{table}

\section{Profile Refinement and Ablation Studies}
\label{app:profile}

\subsection{Profile Refinement}
\label{app:profile-refinement}

\noindent\textbf{Feedback and Controls.}
Each evaluation episode uses 50 PPSD validation users. Ten sampling seeds, 2026 to
2035, determine the feedback and query sets. For each user, 16 query
pairs remain fixed while 24 feedback pairs are sampled. Feedback and query sets
share no image or scene identities. Each recorded choice provides one ordered
preference image pair. The first $N\in\{4,8,12,16,24\}$ pairs of the same feedback
sequence form the reference sets at successive stages. The Frozen Initial Profile
is built from the first four pairs and remains unchanged. Cumulative Feedback
rebuilds the profile from all pairs available at each stage. Reversed New Feedback
swaps the preferred and non-preferred images in newly added pairs while keeping
the initial four pairs unchanged. The primary comparison evaluates Cumulative
Feedback against the Frozen Initial Profile at $N=16$.
Figure~\ref{fig:profile-refinement-main} shows how quality changes as feedback accumulates.

\noindent\textbf{Profile Updates and Inference.}
Profile refinement uses the epoch-91 model trained with seed 2026 and the frozen
LUT autoencoder. All network weights remain frozen. Each profile update re-encodes the retained
ordered preference image pairs and saves a 260-byte Reusable User Profile for later
queries. Reference images and query thumbnails are encoded at 256 and 512 pixels,
respectively. LUTs are applied at full resolution, and metrics are computed at
256 pixels. Predicted edit strength is multiplied by 0.7.

\noindent\textbf{Statistical Evaluation.}
For each feedback stage and condition, raw metric components are first averaged
over queries and all ten feedback sampling seeds within each user. These components
are then averaged equally across the 50 users before computing CQS. Paired 95\%
intervals use 100,000 user-bootstrap resamples, retaining all sampling seeds,
feedback stages, and conditions for each selected user. Each resample repeats
the component aggregation and CQS calculation. The intervals quantify user
variation with model weights fixed. Table~\ref{tab:refinement-metrics} gives the complete paired gains.

\begin{table}[!t]
  \centering
  \caption{Profile refinement on PPSD. $N$ counts accumulated feedback pairs.
  Gains subtract Frozen Initial Profile CQS (4 pairs) from Cumulative Feedback CQS.
  $\dagger$ uses Reversed New Feedback as the comparison at $N=16$.
  Brackets give paired \mbox{95\% user-bootstrap confidence intervals.}}
  \label{tab:refinement-metrics}
  \fontsize{9}{10.5}\selectfont
  \setlength{\tabcolsep}{2.5pt}
  \setlength{\AppMetricWidth}{\dimexpr(\textwidth-.12\textwidth-10\tabcolsep-1\arrayrulewidth)/4\relax}
  \begin{tabular}{L{.12\textwidth}|*{2}{>{\columncolor{AppOrangeA}}w{c}{\AppMetricWidth}>{\columncolor{AppOrangeB}}w{c}{\AppMetricWidth}}}
    \toprule
  & \multicolumn{4}{c}{\cellcolor{AppOrangeHeader}CQS Gain $\uparrow$} \\
  \shortstack[l]{Feedback\\Pairs $N$} & \cellcolor{AppOrangeHeader}$\Delta E_{00}$ & \cellcolor{AppOrangeHeader}LPIPS & \cellcolor{AppOrangeHeader}PSNR & \cellcolor{AppOrangeHeader}SSIM \\
    \midrule
  $8$ & \appgain{\shortstack{$+0.002274$\\$[0.001578, 0.003031]$}} & \appgain{\shortstack{$+0.125910$\\$[0.073599, 0.199634]$}} & \appgain{\shortstack{$+0.289340$\\$[0.196488, 0.379035]$}} & \appgain{\shortstack{$+0.000223$\\$[0.000117, 0.000339]$}} \\[2pt]
  $12$ & \appgain{\shortstack{$+0.003037$\\$[0.002116, 0.004013]$}} & \appgain{\shortstack{$+0.172396$\\$[0.104210, 0.267251]$}} & \appgain{\shortstack{$+0.393652$\\$[0.255991, 0.521696]$}} & \appgain{\shortstack{$+0.000350$\\$[0.000222, 0.000488]$}} \\[2pt]
  $16$ & \appgain{\shortstack{$+0.003215$\\$[0.002309, 0.004167]$}} & \appgain{\shortstack{$+0.176357$\\$[0.107124, 0.271374]$}} & \appgain{\shortstack{$+0.443595$\\$[0.303425, 0.575171]$}} & \appgain{\shortstack{$+0.000359$\\$[0.000224, 0.000503]$}} \\[2pt]
  \cellcolor{AppBlueHeader}$24$ & \appgain{\shortstack{$+0.003531$\\$[0.002684, 0.004431]$}} & \appgain{\shortstack{$+0.184820$\\$[0.114342, 0.282361]$}} & \appgain{\shortstack{$+0.467157$\\$[0.328338, 0.597955]$}} & \appgain{\shortstack{$+0.000374$\\$[0.000237, 0.000520]$}} \\[2pt]
    \midrule
  $16^\dagger$ & \appgain{\shortstack{$+0.003572$\\$[0.002074, 0.005299]$}} & \appgain{\shortstack{$+0.179112$\\$[0.083788, 0.314092]$}} & \appgain{\shortstack{$+0.749985$\\$[0.468387, 1.039789]$}} & \appgain{\shortstack{$+0.000482$\\$[0.000263, 0.000727]$}} \\[2pt]
    \bottomrule
  \end{tabular}
\end{table}

\subsection{Core Ablations}
\label{app:core-ablations}
\label{app:user-contrast}

The full-model control uses a 256-dimensional Reusable User Profile, four learned
pooling tokens, four Transformer blocks, and frozen image encoders. Each structural
ablation changes one component while retaining the other settings. In the Set
Transformer ablation, masked mean pooling replaces the learned set aggregation.
The Explicit Feature Difference and Query Feature ablations set $a_i^+-a_i^-$
and $c_q$ to zero, respectively, while retaining the original input dimensions
and layer shapes.
The variants and full-model control
are initialized from the same personalized PrefLUT checkpoint and trained
for 10 additional epochs on the 471 training users, using seed 2026 and AdamW with
a learning rate of $10^{-5}$. Each run selects its lowest-validation-loss checkpoint:
epoch 3 for the full model, epoch 10 for the feature-difference and Set Transformer
ablations, and epoch 9 for the query-feature ablation. Evaluation uses sampling
seed 2028, 50 users, and 16 reference pairs and 16 query pairs per user. Reference
and query images and metrics use 256 pixels. Profiles use int8 quantization, and
predicted edit strength is multiplied by 0.7. These comparisons measure component
effects under this shared training and evaluation setting, separately from the
epoch-91 model and 512-pixel query inputs with unquantized profiles used for the
main quality results in Table~\ref{tab:personalized-efficiency-main}(A,B). Table~\ref{tab:profile-structure-app}(A) reports absolute CQS and Fixed
Wrong-User Profile PCVP gains for the variants and their shared full-model control.

The Wrong-User Contrast Loss is evaluated in two comparisons using training seeds
2026 and 2027. Within each seed, the full model and the variant without the loss
share the same personalized initialization and undergo 10 additional training
epochs with the same optimization settings as above. Minimum validation loss
selects epoch 5 for both models with seed 2026 and epoch 6 for both with seed 2027.
These runs provide their own full-model controls, separate from the structural
study, and are evaluated at 256 pixels with an edit-strength scale of 0.7.
Seed 2026 combines two evaluation episodes, while seed 2027 uses one.
Each episode includes 50 users and 800 query pairs.

Table~\ref{tab:core-gaps-app}(B) reports each variant's Fixed Wrong-User Profile
PCVP gain minus its matched full model's gain, with paired 95\% user-bootstrap
intervals. Structural ablations use 250,000 resamples, and loss ablations use
100,000. Table~\ref{tab:architecture-ablation-main} uses the same difference
definition and averages the two loss-ablation training runs equally.

\begin{table}[!t]
  \centering
  \caption{Core ablations on PPSD. (A) CQS and Fixed Wrong-User Profile PCVP gains.
  (B) Differences in Fixed Wrong-User Profile PCVP gain, computed as each variant's
  gain minus its matched full model's gain, with paired 95\% user-bootstrap
  confidence intervals. Blue values mark intervals
  entirely below zero. Minus signs denote removed components, with the Set Transformer
  replaced by masked mean pooling.}
  \label{tab:profile-structure-app}
  \label{tab:core-gaps-app}
  \label{tab:user-contrast-app}
  \fontsize{8.5}{10}\selectfont
  \makebox[\textwidth][l]{\textbf{(A) CQS and Wrong-User PCVP Gains}}\par
  \vspace{2pt}
  \setlength{\tabcolsep}{2pt}
  \renewcommand{\arraystretch}{1.12}
  \setlength{\AppMetricWidth}{\dimexpr(\textwidth-.22\textwidth-18\tabcolsep-2\arrayrulewidth)/8\relax}
  \begin{tabular}{L{.22\textwidth}|*{2}{>{\columncolor{AppOrangeA}}w{r}{\AppMetricWidth}>{\columncolor{AppOrangeB}}w{r}{\AppMetricWidth}}|*{2}{>{\columncolor{AppBlueA}}w{r}{\AppMetricWidth}>{\columncolor{AppBlueB}}w{r}{\AppMetricWidth}}}
    \toprule
   & \multicolumn{4}{c|}{\cellcolor{AppOrangeHeader}Correct User Profile CQS $\uparrow$}
   & \multicolumn{4}{c}{\cellcolor{AppBlueHeader}Wrong-User PCVP Gain $\uparrow$} \\
  Variant & $\Delta E_{00}$ & LPIPS & PSNR & SSIM & $\Delta E_{00}$ & LPIPS & PSNR & SSIM \\
    \midrule
  \shortstack[l]{$-$ Explicit Feature\\\phantom{$-$ }Difference}
  & .20928 & 21.00842 & 31.73068 & .91467 & +.00102 & +.10579 & +.10323 & +.00028 \\
  $-$ Set Transformer
  & .19050 & 19.72441 & 29.69401 & .91261 & -.00014 & -.00014 & -.00290 & -.00001 \\
  $-$ Query Feature
  & .20657 & 19.77052 & 33.42625 & .91171 & +.00081 & +.08989 & +.04642 & +.00028 \\
    \midrule
  \cellcolor{AppBlueHeader}PrefLUT
  & .21034 & 21.20822 & 31.94439 & .91523 & +.00410 & +.42875 & +.38935 & +.00139 \\
    \bottomrule
  \end{tabular}
  \par\vspace{6pt}
  \makebox[\textwidth][l]{\textbf{(B) Change in Wrong-User PCVP Gain (Variant Minus Full Model)}}\par
  \vspace{2pt}
  \fontsize{7.5}{9}\selectfont
  \setlength{\tabcolsep}{2pt}
  \renewcommand{\arraystretch}{1.12}
  \setlength{\AppMetricWidth}{\dimexpr(\textwidth-.20\textwidth-10\tabcolsep-1\arrayrulewidth)/4\relax}
  \begin{tabular}{L{.20\textwidth}|*{2}{>{\columncolor{AppBlueA}}w{c}{\AppMetricWidth}>{\columncolor{AppBlueB}}w{c}{\AppMetricWidth}}}
    \toprule
  Variant & \cellcolor{AppBlueHeader}\shortstack{$\Delta E_{00}$\\$(10^{-3})$} & \cellcolor{AppBlueHeader}LPIPS & \cellcolor{AppBlueHeader}PSNR & \cellcolor{AppBlueHeader}\shortstack{SSIM\\$(10^{-3})$} \\
    \midrule
  \rowcolor{AppBlueHeader}\multicolumn{5}{l}{Structural ablations} \\
  \shortstack[l]{$-$ Explicit Feature\\\phantom{$-$ }Difference}
  & \appgain{\shortstack{$-3.074$\\$[-5.035,-1.074]$}}
  & \appgain{\shortstack{$-0.322957$\\$[-.533102,-.168424]$}}
  & \shortstack{$-0.286126$\\$[-.619128,.028585]$}
  & \appgain{\shortstack{$-1.116$\\$[-1.782,-0.527]$}} \\[2pt]
  $-$ Set Transformer
  & \appgain{\shortstack{$-4.238$\\$[-6.382,-2.060]$}}
  & \appgain{\shortstack{$-0.428884$\\$[-.680801,-.245940]$}}
  & \appgain{\shortstack{$-0.392256$\\$[-.749216,-.053438]$}}
  & \appgain{\shortstack{$-1.401$\\$[-2.165,-0.735]$}} \\[2pt]
  $-$ Query Feature
  & \appgain{\shortstack{$-3.290$\\$[-5.462,-1.312]$}}
  & \appgain{\shortstack{$-0.338858$\\$[-.566963,-.175087]$}}
  & \shortstack{$-0.342936$\\$[-.855309,.138272]$}
  & \appgain{\shortstack{$-1.110$\\$[-1.758,-0.539]$}} \\[2pt]
    \midrule
  \rowcolor{AppOrangeHeader}\multicolumn{5}{l}{Loss ablations} \\
  \shortstack[l]{$-$ Wrong-User Contrast\\\phantom{$-$ }Loss (Seed 2026)}
  & \appgain{\shortstack{$-0.037053$\\$[-.065623,-.007432]$}}
  & \appgain{\shortstack{$-0.004843$\\$[-.007850,-.002235]$}}
  & \shortstack{$-0.002883$\\$[-.006968,.001180]$}
  & \appgain{\shortstack{$-0.013593$\\$[-.024760,-.004833]$}} \\[2pt]
  \shortstack[l]{$-$ Wrong-User Contrast\\\phantom{$-$ }Loss (Seed 2027)}
  & \appgain{\shortstack{$-0.035488$\\$[-.064323,-.005684]$}}
  & \appgain{\shortstack{$-0.004894$\\$[-.009005,-.001913]$}}
  & \shortstack{$-0.001582$\\$[-.006312,.003404]$}
  & \appgain{\shortstack{$-0.013424$\\$[-.027026,-.001900]$}} \\[2pt]
    \bottomrule
  \end{tabular}
\end{table}

Removing learned set aggregation reduces the Fixed Wrong-User Profile PCVP gain,
with paired intervals entirely below zero on all four metrics. Removing the
Explicit Feature Difference, Query Feature, or Wrong-User Contrast Loss gives
intervals entirely below zero on three metrics. Each comparison uses the full model
trained under the same settings as the corresponding ablation.

\subsection{Profile Width}
\label{app:profile-width}

Table~\ref{tab:profile-dimension-app} compares profile widths using the initialization,
10-epoch training budget, checkpoint-selection rule, and evaluation settings of
the structural study in Appendix~\ref{app:core-ablations}. The 256-dimensional
reference is the same full-model checkpoint used in Table~\ref{tab:profile-structure-app}(A).
The selected epochs for widths 32, 64, 128, 256, and 512 are 10, 9, 3, 3, and 3,
respectively. An int8 profile stores one byte per dimension and a four-byte scale.
For each width, the paired CQS difference is measured against the 256-dimensional
profile. Quality is considered retained when the lower bound of the 95\% interval
exceeds $-1\%$ of the reference CQS on every metric. Among the tested widths,
the 256-dimensional profile is the smallest that meets this quality criterion,
using 260 bytes.
Increasing the width to 512 provides comparable quality while nearly doubling storage.

\begin{table}[!htbp]
  \centering
  \caption{Reusable User Profile dimension on PPSD. CQS uses the correct user
  profile, and PCVP gain compares it with the Fixed Wrong-User Profile.
  \mbox{The standard profile has 256 dimensions.}}
  \label{tab:profile-dimension-app}
  \fontsize{8.5}{10}\selectfont
  \setlength{\tabcolsep}{2pt}
  \renewcommand{\arraystretch}{1.12}
  \setlength{\AppMetricWidth}{\dimexpr(\textwidth-.12\textwidth-18\tabcolsep-2\arrayrulewidth)/8\relax}
  \begin{tabular}{L{.12\textwidth}|*{2}{>{\columncolor{AppOrangeA}}w{r}{\AppMetricWidth}>{\columncolor{AppOrangeB}}w{r}{\AppMetricWidth}}|*{2}{>{\columncolor{AppBlueA}}w{r}{\AppMetricWidth}>{\columncolor{AppBlueB}}w{r}{\AppMetricWidth}}}
    \toprule
   & \multicolumn{4}{c|}{\cellcolor{AppOrangeHeader}Correct User Profile CQS $\uparrow$}
   & \multicolumn{4}{c}{\cellcolor{AppBlueHeader}Wrong-User PCVP Gain $\uparrow$} \\
  \shortstack[l]{Profile\\Dimension} & $\Delta E_{00}$ & LPIPS & PSNR & SSIM & $\Delta E_{00}$ & LPIPS & PSNR & SSIM \\
    \midrule
  32 & .20557 & 20.8993 & 31.2043 & .91469 & .00138 & .1432 & .1089 & .00046 \\
  64 & .20773 & 21.0845 & 31.4524 & .91501 & .00237 & .2263 & .2219 & .00072 \\
  128 & .20806 & 21.0872 & 31.5174 & .91502 & .00312 & .3105 & .2948 & .00099 \\
  \cellcolor{AppBlueHeader}256 & .21034 & 21.2082 & 31.9444 & .91523 & .00410 & .4287 & .3894 & .00139 \\
  512 & .21014 & 21.1969 & 31.9005 & .91520 & .00411 & .4284 & .3896 & .00138 \\
    \bottomrule
  \end{tabular}
\end{table}

\section{Limitations}
\label{app:limitations}

PPSD does not release its official user split or episode generator, so this evaluation
uses a deterministic split and sampling procedure. Validation users are excluded
from gradient-based training but used for checkpoint selection, including selection
of epoch 91 by CQS on the reported episode. These validation-set estimates may
therefore be optimistic. Repeated episodes measure sensitivity to reference and
query sampling. Independent-user generalization requires users excluded from model
development and selection. Standard PrefLUT uses a global 3D LUT. Spatially varying
retouching requires the additional spatial component used in the FiveK and PPR10K
configurations (Appendix~\ref{app:task-settings}). PPSD tests personalized conditioning.
FiveK and PPR10K evaluate enhancement toward fixed expert targets.

\section{Extended Qualitative Evaluation}
\label{app:qualitative}

\begin{figure*}[!ht]
  \centering
  \includegraphics[width=\textwidth]{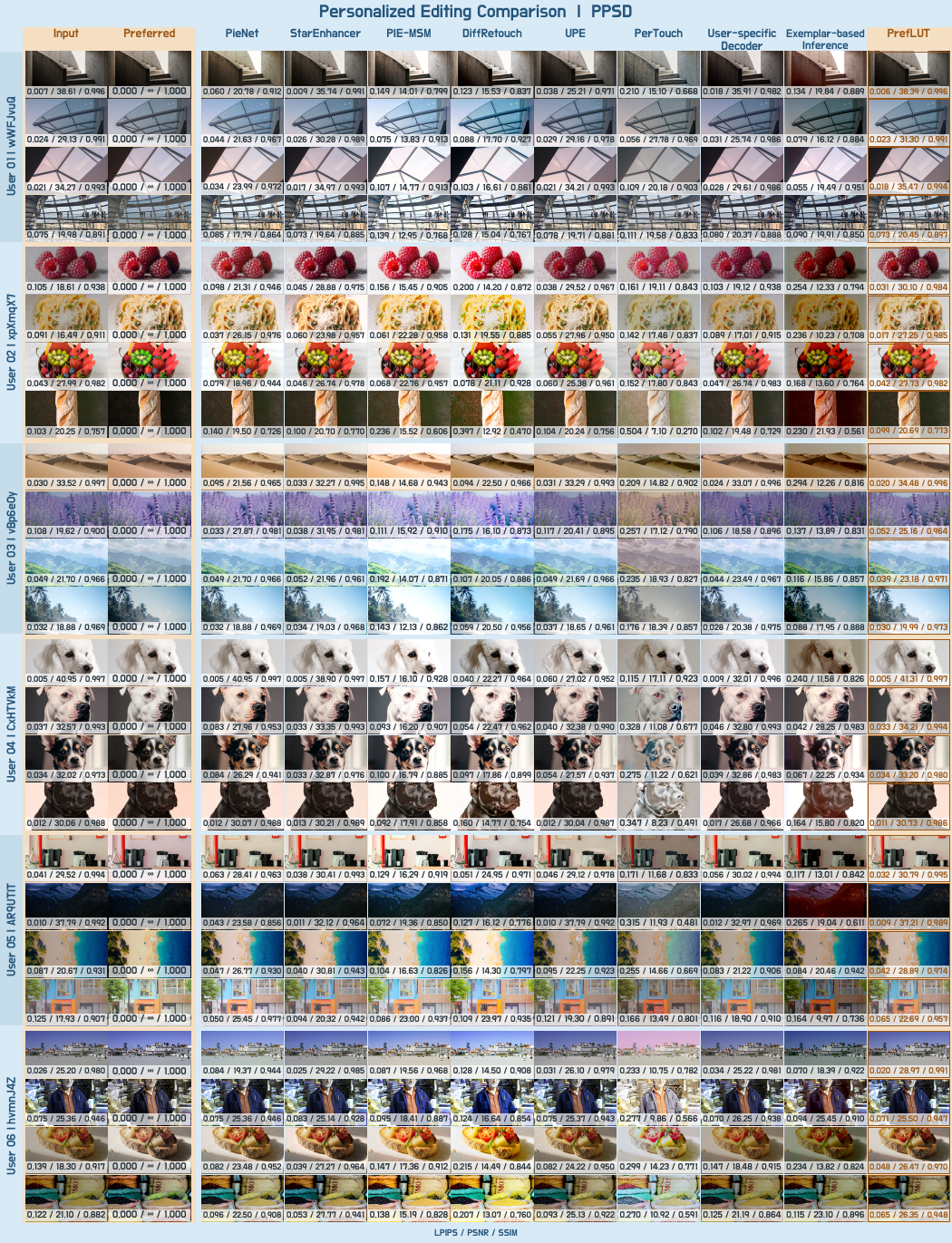}
  \caption{\textbf{Personalized editing comparisons on PPSD.} Queries are grouped
  by user. Input and Preferred show the non-preferred query and its paired preferred
  target. All methods share queries and 16 reference pairs per user.
  Labels report direct LPIPS $\downarrow$/PSNR $\uparrow$/SSIM $\uparrow$ against Preferred.}
  \label{fig:qualitative-comparison-app}
\end{figure*}

\noindent\textbf{Personalized Editing.}\quad
Figure~\ref{fig:qualitative-comparison-app} compares methods on the same queries and
reference sets, grouped by user. Preferred targets include both brighter and darker
appearances, so the desired edit direction varies across examples. PrefLUT adjusts
brightness and color toward these targets across animal, food, and outdoor scenes.
The side-by-side outputs also show differences in saturation and preservation of
scene detail.

\begin{figure*}[p]
  \centering
  \includegraphics[width=\textwidth,height=0.88\textheight,keepaspectratio]{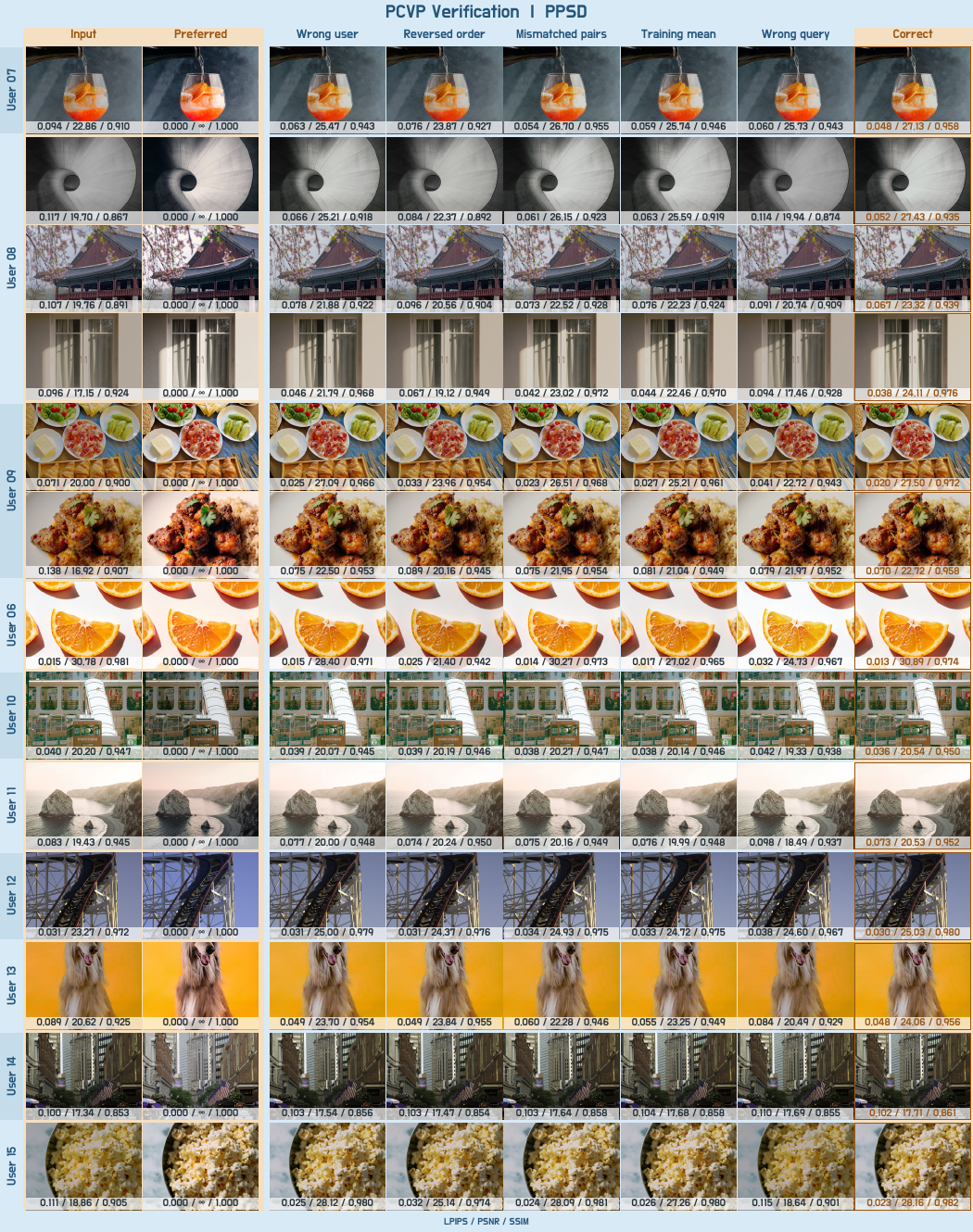}
  \caption{\textbf{PrefLUT PCVP examples on PPSD.} Each row compares Correct
  conditioning with the five controls in Appendix~\ref{app:pcvp-details}.
  Wrong query denotes the Cyclic Wrong-Query Control, which uses another query from
  the same user to predict the LUT and edit strength applied to the displayed input. Labels report direct LPIPS $\downarrow$/PSNR $\uparrow$/SSIM
  $\uparrow$ against Preferred.}
  \label{fig:qualitative-pcvp-app}
\end{figure*}

\noindent\textbf{PCVP.}\quad
Figure~\ref{fig:qualitative-pcvp-app} holds the displayed input fixed and changes only
the profile or query condition. The tunnel, food, and building examples show visible
changes in brightness and color balance under the five controls. Correct conditioning
more closely follows the paired preferred appearance. These comparisons connect the
PCVP tests to visible editing behavior on individual queries.

\begin{figure*}[p]
  \centering
  \includegraphics[width=\textwidth,height=0.88\textheight,keepaspectratio]{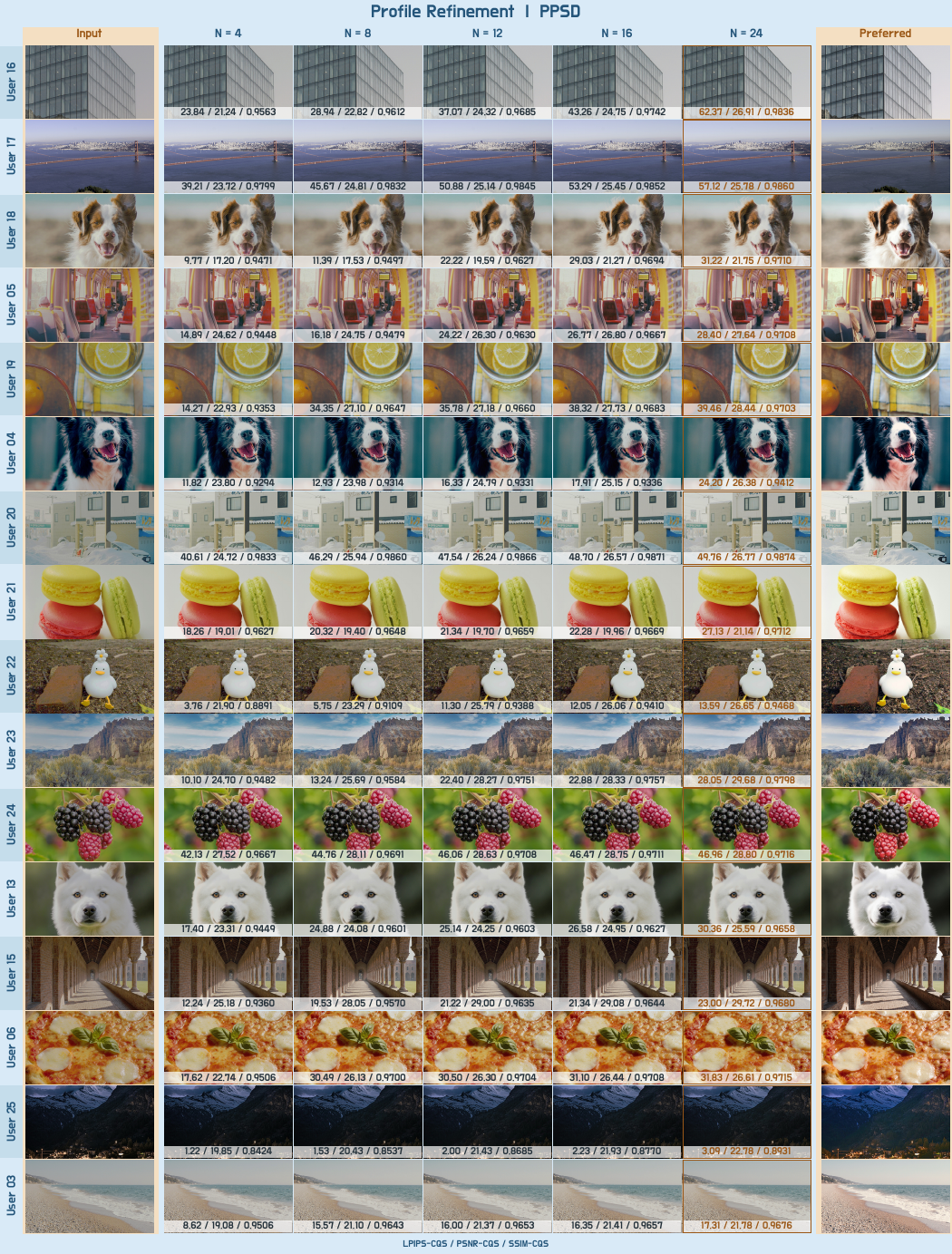}
  \caption{\textbf{PrefLUT profile refinement on PPSD.} Each row fixes the query
  and network weights while the profile is rebuilt from the first $N=4,8,12,16,24$
  feedback pairs. Input is the non-preferred query, and Preferred is its paired
  target. Labels report per-output \mbox{LPIPS-CQS}, \mbox{PSNR-CQS},
  and \mbox{SSIM-CQS} (all $\uparrow$). Per-output CQS compares each output with
  Preferred and \mbox{Input (Appendix~\ref{app:metric-statistics}).}}
  \label{fig:qualitative-refinement-app}
\end{figure*}

\noindent\textbf{Profile Refinement.}\quad
Figure~\ref{fig:qualitative-refinement-app} follows each query as feedback accumulates
from $N=4$ to $24$ pairs. Additional feedback refines brightness, saturation, and
color balance toward the paired target while the scene layout is retained. The query
and network weights stay fixed, so the changes result from updating the Reusable
User Profile. Per-output CQS labels summarize the combined changes in fidelity and
preference alignment alongside the visual results.

\clearpage
\clearpage
\clearpage

\end{document}